\documentclass[sigconf]{acmart}

\usepackage{algorithm}
\usepackage{algorithmic}
\usepackage{newfloat}
\usepackage{listings}

\usepackage{colortbl}
\usepackage{graphicx}
\usepackage{amsmath}
\usepackage{amsthm}

\usepackage{multirow}
\usepackage{csquotes}
\usepackage{subcaption}
\usepackage{caption}
\usepackage{adjustbox}
\usepackage{booktabs}
\usepackage[toc,page,header]{appendix}
\usepackage{makecell}
\usepackage{booktabs}

\usepackage[most]{tcolorbox}
\usepackage{listings}
\usepackage{float}
\usepackage[x11names]{xcolor} 
\usepackage{pgf}
\usepackage{pgfplots}

\definecolor{backgroundblue}{HTML}{61CBF4} 
\definecolor{darkblue}{RGB}{0, 0, 139}
\definecolor{darkgreen}{rgb}{0.0, 0.5, 0.0}
\definecolor{orchid}{rgb}{0.85, 0.44, 0.84}
\definecolor{mediumorchid}{rgb}{0.73, 0.33, 0.83}

\theoremstyle{definition}
\newtheorem{definition}{Definition}[section]

\theoremstyle{remark}

\usepackage{tcolorbox}
\tcbset{colback=white}
\colorlet{titleblue}{blue!80!black}
\colorlet{titlered}{red!80!black}
\colorlet{titlegreen}{green!80!black}
\colorlet{darkgreen}{green!50!black}

\newcommand{\TBlack}[1]{\textnormal{\ttfamily\color{black}#1}\unskip}

\usepackage{pgf}
\usepackage{pgfplots}
\usepackage{dsfont}
\usepackage{graphicx}
\usepackage[dvipsnames,svgnames,table]{xcolor}

\pgfplotsset{compat=1.10}
\usepackage{tabularx} 
\usepackage{collcell}
\usepackage{xfp}

\definecolor{goodgreen}{HTML}{c5eecc}
\definecolor{goodred}{HTML}{ffc7ce}
\newcommand{\MinNumber}{0}%
\newcommand{\MaxNumber}{100}%

\newcommand{\ApplyGradient}[1]{%
    \begingroup
    \let\relax\empty
    \if\relax\detokenize{#1}\relax\else
        
            \newcommand\Percent{\fpeval{100.0*(#1-\MinNumber)/(\MaxNumber-\MinNumber)}}
            \pgfmathsetmacro{\PercentColor}{\Percent}
            \xdef\PercentColor{\PercentColor}%
            \cellcolor{goodgreen!\PercentColor!goodred}
    \fi

    \endgroup
    {#1}
}

\newcolumntype{H}[2]{>{\SetGradientLimits{#1}{#2}\collectcell\ApplyGradient}X<{\endcollectcell}}
\tcbset{
  aibox/.style={
    width=\linewidth,
    top=10pt,
    colback=white,
    colframe=black,
    colbacktitle=black,
    enhanced,
    center,
    attach boxed title to top left={yshift=-0.1in,xshift=0.15in},
    boxed title style={boxrule=0pt,colframe=white,},
  }
}
\newtcolorbox{AIbox}[2][]{aibox,title=#2,#1}
\newcommand{\applyGradient}[1]{%
    \ifdim #1 pt > 80pt \cellcolor{black!3}{#1}%
    \else\ifdim #1 pt > 60pt \cellcolor{black!7}{#1}%
    \else\ifdim #1 pt > 40pt \cellcolor{black!12}{#1}%
    \else\ifdim #1 pt > 20pt \cellcolor{black!18}{#1}%
    \else\cellcolor{black!25}{#1}%
    \fi\fi\fi\fi%
}

\AtBeginDocument{%
  }

\copyrightyear{2026}
\acmYear{2026}
\setcopyright{cc}
\setcctype{by}
\acmConference[KDD '26]{Proceedings of the 32nd ACM SIGKDD Conference on Knowledge Discovery and Data Mining V.1}{August 09--13, 2026}{Jeju Island, Republic of Korea}
\acmBooktitle{Proceedings of the 32nd ACM SIGKDD Conference on Knowledge Discovery and Data Mining V.1 (KDD '26), August 09--13, 2026, Jeju Island, Republic of Korea}
\acmPrice{}
\acmDOI{10.1145/3770854.3785678}
\acmISBN{979-8-4007-2258-5/2026/08}

\acmSubmissionID{46}

\begin{document}

\title{MM-SpuBench: Towards Better Understanding of Spurious Biases in Multimodal LLMs}

\author{Wenqian Ye}
\affiliation{%
  \institution{University of Virginia}
  \city{Charlottesville}
  \state{VA}
  \country{USA}}
\email{wenqian@virginia.edu}

\author{Bohan Liu}
\affiliation{%
  \institution{University of Virginia}
  \city{Charlottesville}
  \state{VA}
  \country{USA}}
\email{qzp4ta@virginia.edu}

\author{Guangtao Zheng}
\affiliation{%
  \institution{University of Virginia}
  \city{Charlottesville}
  \state{VA}
  \country{USA}}
\email{gz5hp@virginia.edu}

\author{Di Wang}
\affiliation{%
  \institution{University of Virginia}
  \city{Charlottesville}
  \state{VA}
  \country{USA}}
\email{azm7tq@virginia.edu}

\author{Yunsheng Ma}
\affiliation{%
  \institution{Purdue University}
  \city{West Lafayette}
  \state{IN}
  \country{USA}}
\email{yunsheng@purdue.edu}

\author{Xu Cao}
\affiliation{%
  \institution{University of Illinois Urbana-Champaign}
  \city{Champaign}
  \state{IL}
  \country{USA}}
\email{xucao2@illinois.edu}

\author{Bolin Lai}
\affiliation{%
  \institution{Georgia Institute of Technology}
  \city{Atlanta}
  \state{GA}
  \country{USA}}
\email{bolin.lai@gatech.edu}

\author{James M. Rehg}
\affiliation{%
  \institution{University of Illinois Urbana-Champaign}
  \city{Champaign}
  \state{IL}
  \country{USA}}
\email{jrehg2@illinois.edu}

\author{Aidong Zhang}
\affiliation{%
  \institution{University of Virginia}
  \city{Charlottesville}
  \state{VA}
  \country{USA}}
\email{aidong@virginia.edu}

\renewcommand{\shortauthors}{Ye et al.}

\begin{abstract}

Spurious bias, a tendency to exploit spurious correlations between superficial input attributes and prediction targets, has revealed a severe robustness pitfall in classical machine learning problems. Multimodal Large Language Models (MLLMs), which leverage pretrained vision and language models, have recently demonstrated strong capability in joint vision-language understanding. However, both the presence and severity of spurious biases in MLLMs remain poorly understood. In this work, we address this gap by analyzing the spurious biases in the multimodal setting and uncovering the specific inference-time data patterns that can manifest this problem. To support this analysis, we introduce \textsc{MM-SpuBench}, a comprehensive, human-verified benchmark dataset consisting of image-class pairs annotated with core and spurious attributes, grounded in our taxonomy of nine distinct types of spurious correlations. The benchmark is constructed using human-interpretable attribute information to capture a wide range of spurious patterns reflective of real-world knowledge. Leveraging this benchmark, we conduct a comprehensive evaluation of the state-of-the-art open-source and proprietary MLLMs with both standard accuracy and the proposed Conditional Generation Likelihood Advantage (CGLA). Our findings highlight the persistence of reliance on spurious correlations and the difficulty of mitigation on our benchmark. We hope this work can inspire new technical strides to mitigate these biases. Our benchmark is publicly available at \url{https://huggingface.co/datasets/mmbench/MM-SpuBench}.

\end{abstract}



\maketitle

\section{Introduction}


Recently, we have witnessed the rise of highly performant Large Language Models (LLMs)~\cite{chowdhery2023palm,chung2024scaling,devlin2018bert,touvron2023llama,wei2021finetuned,zhang2022opt} and Vision Foundation Models (VFMs)~\cite{oquab2023dinov2,he2022masked} powered by advancements in vision language modeling, as well as the availability of large-scale training data and substantial computational resources. Building on these advancements, multimodal Large Language Models (MLLMs) ~\cite{mckinzie2024mm1,achiam2023gpt,claude3_family,team2023gemini,dai2024instructblip, liu2024visual, meta2024llama32, zhu2025internvl3exploringadvancedtraining} emerge as the new frontier of foundation models by integrating both LLMs and VFMs for joint visual and text understanding. MLLMs have demonstrated significant performance gains in visual reasoning tasks, such as image perception~\cite{lu2024seeing,huang2023language}, visual question answering~\cite{yin2024lamm}, and instruction following~\cite{bitton2023visit}.

\begin{figure}[t]
  \centering
  \includegraphics[width=0.46\textwidth]{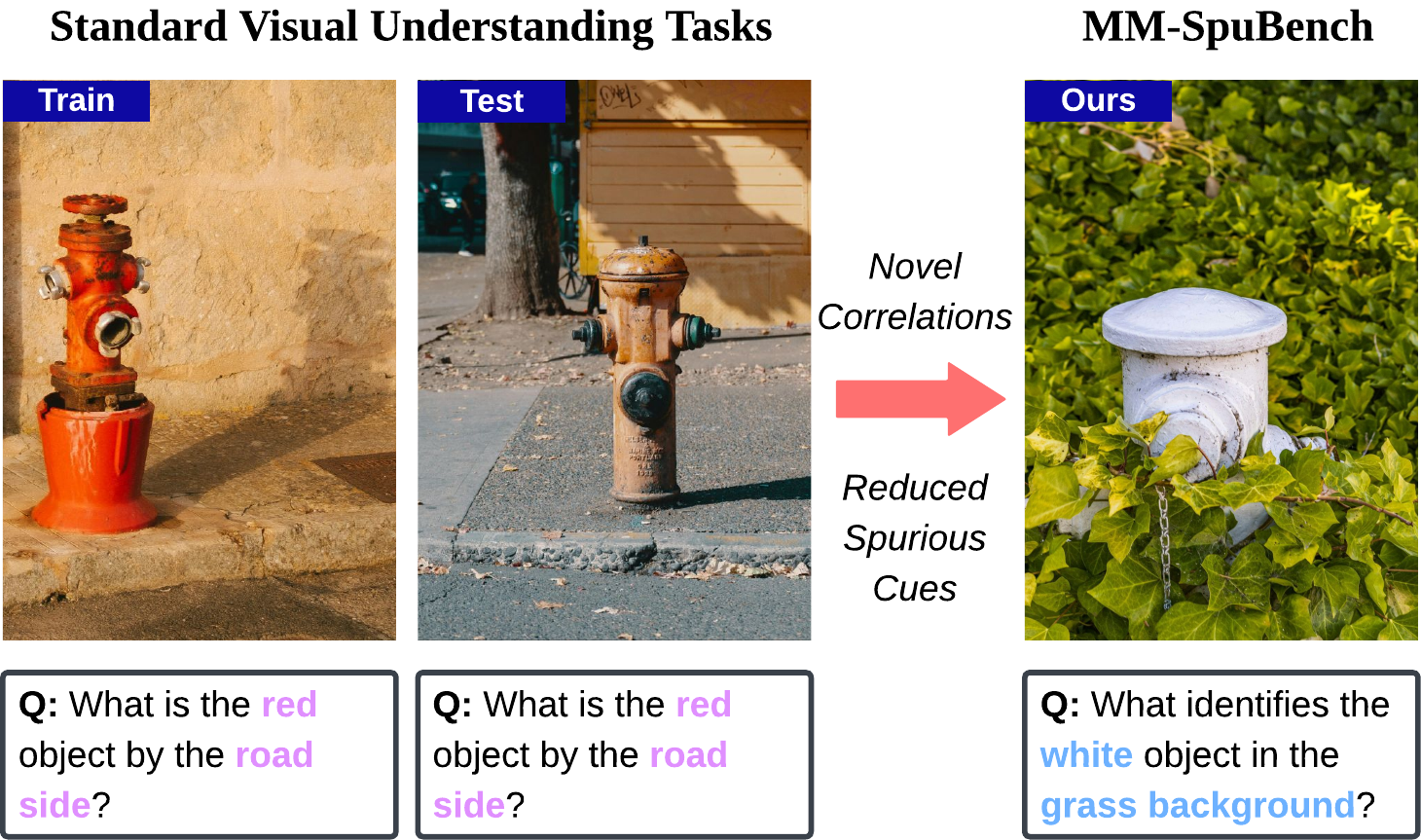}
  \caption{Schematic illustration of standard visual understanding tasks and \textsc{MM-SpuBench}. In standard tasks, both training and testing data consist of spurious correlations, such as fire hydrants frequently co-occurring with red color and roadside. In \textsc{MM-SpuBench}, we introduce novel correlations by reducing spurious cues, such as fire hydrants paired with white color and grass background.}
  \label{fig:schematic}
\end{figure}


\begin{figure*}[!t]
    \centering
    \includegraphics[width=0.88\linewidth]{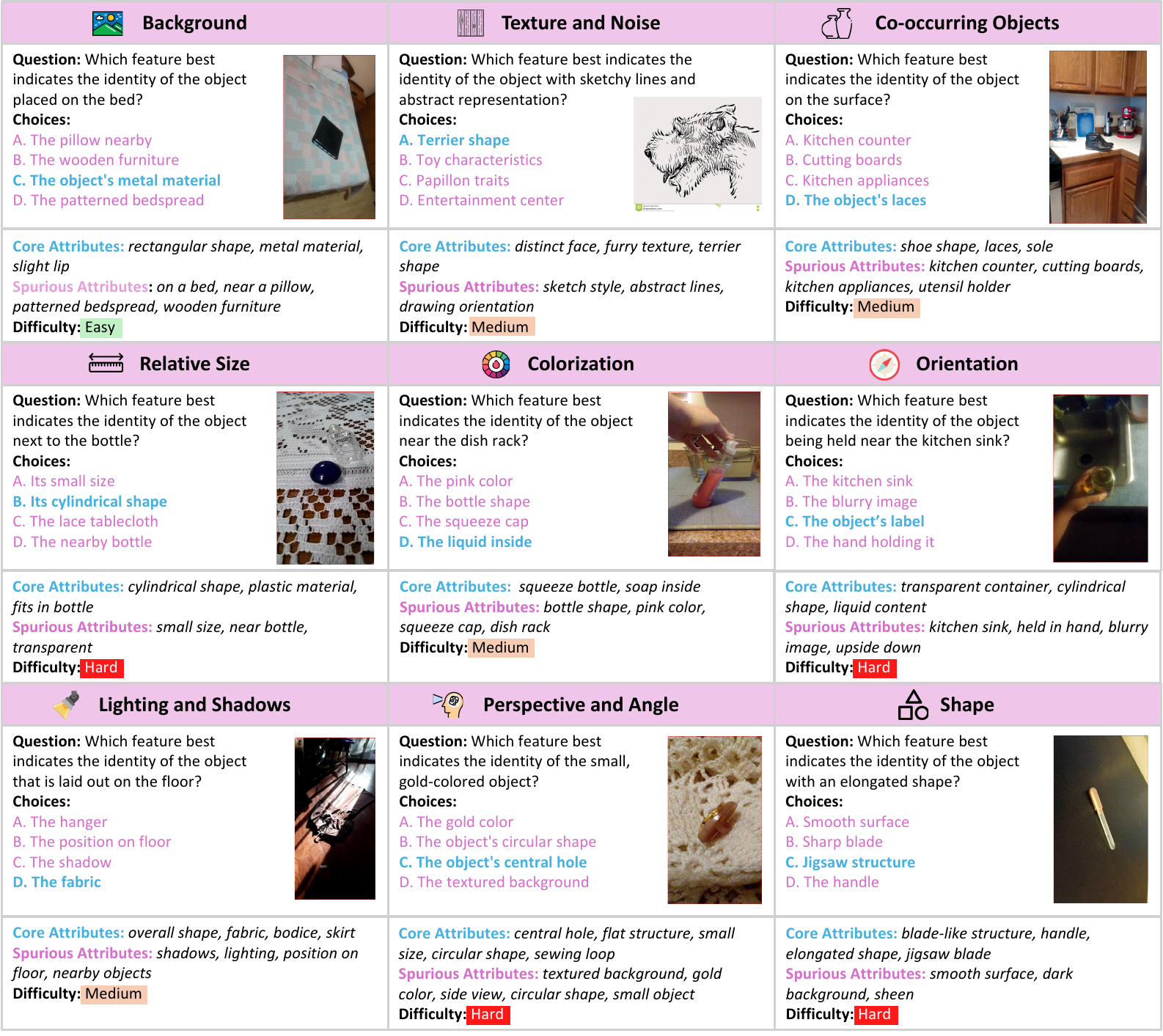}
    \caption{
    Samples from each spurious correlation type in \textsc{MM-SpuBench}. The images and questions require human-level cognitive abilities to judge the core information in the provided choices.
    }
    \label{fig:illustration}
\end{figure*}

Despite the impressive performance of MLLMs, the robustness of MLLMs remains largely under-explored. One critical robustness issue in deep learning models is the \textit{spurious bias}, a tendency to exploit spurious correlations between superficial input attributes and prediction target \cite{ye2024spurious} for higher in-distribution performance. For example, image classifiers tend to identify a \texttt{waterbird} by solely using the \texttt{water background} (\textit{spurious attribute}) that frequently co-occurs with it in the training data, whereas the bird’s intrinsic features (\textit{core attributes}), such as color and shape, should be the true indicators for accurate predictions \cite{geirhos2020shortcut,sagawa2019distributionally}. Previous research works have primarily defined and mitigated spurious bias in single-modality classification tasks \cite{sagawa2019distributionally,liu2021just,nam2022spread,kirichenko2022last,zheng2024learning}. 

Getting back to MLLMs, their pre-trained vision and text encoders also tend to learn spurious correlations within their respective modalities~\cite{ghosal2024vision,yang2023mitigating,eisenstein2022informativeness,wang2022identifying}. In the MLLM training for visual grounding, the ability to align specific visual regions with corresponding textual labels may be hindered by spurious correlations in both vision and language modalities. Consequently, at the inference stage, where the captured multi-modal spurious correlations break due to distribution shifts, MLLMs tend to make biased predictions. With the main object of \texttt{the fire hydrant} shown in Fig.~\ref{fig:schematic}, multimodal spurious correlations in the training (i.e., \texttt{fire hydrant} and \texttt{red color}) no longer hold in the inference data (i.e., \texttt{fire hydrant} and \texttt{white color}), leading the model to select the wrong answer in the VQA inference data. We refer to the above phenomenon as \textit{multimodal spurious bias}. Given the prevalence of spurious biases in classical machine learning problems, a natural question arises:

\begin{center}
\begin{tcolorbox}[
  width=0.85\linewidth,
  colback=gray!5,
  colframe=black!20,
  boxrule=0.5pt,
  arc=2mm,
  left=6pt,
  right=6pt,
  top=2pt,
  bottom=2pt
]
\centering
\textit{Do MLLMs also exhibit spurious biases?\\
If so, to what extent are they affected?}
\end{tcolorbox}
\end{center}

In this paper, we formally define the concept of \textit{multimodal spurious bias} that involves the spurious correlations within multiple modalities. Our definition begins with the notion of an \textit{anchor}, an abstract, semantically meaningful concept that is central to a VQA task and is grounded in both modalities. For example, the anchor for the VQA from the training data in Fig.~\ref{fig:schematic} is the concept of \texttt{fire hydrant}, and it exists in both the image and the text. The introduction of \textit{anchor} allows us to define spurious correlations within vision and language modalities conveniently. There are spurious correlations between visual features ``fire hydrant'' and ``roadside'' and the spurious correlation between text ``fire hydrant'' and ``red'' or ``road'', as shown in Fig.~\ref{fig:schematic}. These correlations occur frequently in the training data and when such correlations break (i.e., with \texttt{white color} and \texttt{grass background}), the MLLMs will lead to failures in prediction.

To better study and quantify the multimodal spurious bias, it is essential to construct dedicated evaluation data that targets robustness failures in MLLMs. To address this need, we propose \textsc{MM-SpuBench}, a comprehensive VQA benchmark with nine categories of spurious correlations specifically designed to evaluate the reliance of MLLMs on instance-level spurious correlations learned during training. To curate the dataset, we introduce a simple yet effective semiautomatic pipeline that leverages parametric knowledge \cite{achiam2023gpt} to extract and annotate attributes of inference data. We carefully select 10,773 image samples from five open-source datasets, design 2,400 VQA questions containing derived core/spurious attributes, and categorize spurious biases into 9 categories, shown in Fig. \ref{fig:illustration}. We then propose both Standard VQA Accuracy within the nine categories and a novel Conditional Generation Likelihood Advantage (CGLA) on \textsc{MM-SpuBench} as metrics for evaluating fine-grained spurious biases. CGLA measures the impact of these attributes on the output token distributions of MLLMs. These two evaluation metrics complement each other by providing \textit{structured} and \textit{unstructured} evaluation setups and offer both top-down and bottom-up studies on examining spurious correlations in MLLMs. With the novel correlations and reduced spurious correlations in our benchmark, models rely minimally on learned spurious correlations. Therefore, their tendency towards spurious biases can be measured quantitatively.

\textbf{Our contributions are summarized as follows:} (1). We formally define multimodal spurious bias in MLLMs, highlighting how spurious correlations can compromise robustness and lead to failures in current MLLMs. (2). We propose \textsc{MM-SpuBench}, a novel benchmark featuring 10,773 realistic images with concept-based attribute information, paired with a set of 2,400 VQA data, designed to systematically evaluate current MLLMs across 9 distinct categories of spurious biases. (3). We conduct an in-depth analysis of current representative MLLMs with two proposed metrics and perform experiments on various prompting techniques to mitigate spurious biases. The results reveal some current limitations and shed light on future research directions.

\section{Related Works}

\paragraph{Robustness in multimodal LLMs.}
Recent proprietary MLLMs, such as GPT-4V~\cite{achiam2023gpt}, Claude~\cite{claude3_family}, and Gemini~\cite{team2023gemini}, have demonstrated notable robustness to various distribution shifts, showcasing the potential in handling diverse and challenging real-world scenarios. Moreover, thanks to the high-quality visual instruction tuning data, we have seen improved robustness~\cite{tong2024eyes} in open-source models like InternVL3~\cite{zhu2025internvl3exploringadvancedtraining}, Llama-3.2-Vision~\cite{meta2024llama32}, and LLaVA~\cite{liu2024visual}. Nevertheless, MLLMs still face challenges in handling visually complex images with spurious correlations, which cause hallucinations and non-trustworthy behaviors~\cite{gavrikov2024vision, han2024instinctive}, exposing the limitations in visual search mechanisms~\cite{wu2023textit,barbany2024leveraging} and visual grounding capabilities~\cite{tong2024eyes,rasheed2024glamm,zhang2025llava}. Our paper focuses on the fundamental spurious bias issue in the multimodal setting, as it reflects a broad family of biases prevalent in current MLLMs. 

\paragraph{Spurious attribute detection.} 
Spurious attributes can negatively impact a model's generalization capabilities \cite{geirhos2020shortcut,yongspurious} and are commonly used to reveal the model's robustness pitfalls. Some previous studies predefine \cite{luo2023zero,qiu2020semanticadv,wu2023discover} a list of spurious attributes relevant to the task, which also refer to group labels. However, such labels require domain knowledge \cite{clark2019don,nauta2021uncovering} and extensive human annotation efforts \cite{nushi2018towards,zhang2018manifold}. For example, object backgrounds \cite{xiao2021noise} and image textures \cite{geirhos2018imagenettrained} have been identified as spurious attributes that can bias the predictions of deep learning models. Recent research has employed explainable data generation methods \cite{plumb2022finding,abid2022meaningfully} to detect or exploit spurious biases in LLMs \cite{prabhu2023lance,hosseini2025seeing,zheng2024spuriousness,ye2025sage}. In our work, we use the LLM's parametric knowledge to parse the information in ground truth labels and incorrectly predicted labels by state-of-the-art vision encoders for spurious concepts that are human-understandable. These concepts are used to build challenging VQA tasks for benchmarking MLLMs.

\paragraph{Benchmarks on multimodal LLMs.}
In classical machine learning problems, there have been several benchmarks \cite{sagawa2019distributionally,zheng2024benchmarking} to evaluate spurious correlations on classification tasks. In the multimodal setting, previous benchmarks such as TextVQA~\cite{singh2019towards} and GQA~\cite{hudson2019gqa} have focused on traditional VQA queries. More recently, works such as MM-Vet~\cite{yu2023mm}, POPE~\cite{li2023evaluating}, and MM-Bench~\cite{liu2023mmbench} have been developed to specifically evaluate multimodal LLMs in terms of hallucination, reasoning, and robustness. These evaluations have uncovered that multimodal LLMs can suffer from hallucination~\cite{huang2023opera,chen2024mllm,bai2024hallucination}, catastrophic forgetting~\cite{zhai2024investigating,zhu2024model}, and a lack of robustness~\cite{wang2024stop,cui2024robustness,biswas2024robustness}. Unlike previous VQA benchmarks, which only include question-answer data, our benchmark also incorporates concept-based information on both core and spurious attributes. This addition helps researchers distinguish between core and spurious information, thereby facilitating the future development of spurious bias mitigation methods.


\section{Spurious Biases in Multimodal LLMs}
\label{sec:sb}

\subsection{Preliminary}
We consider a typical multimodal setting involving a vision modality $\mathcal{X}$ and a language modality $\mathcal{Y}$. We introduce an abstract concept \textit{anchor} $a$ for each image-text encoding pair $(\mathbf{x}, \mathbf{y})$, representing the primary entity or object that forms the central focus. Given an image input $\mathbf{x} \in \mathcal{X}$ and a text input $\mathbf{y} \in \mathcal{Y}$, an MLLM algorithm learns a mapping function $\phi:\mathcal{X} \times \mathcal{Y} \rightarrow \mathcal{C}$, such that the response $c = \phi(\mathbf{x}, \mathbf{y} | a)$, where $c \in \mathcal{C} \subset \mathcal{Y}$ is generated autoregressively based on the image $\mathbf{x}$ and text context $\mathbf{y}$. In the VQA setting, the \textit{anchor} label $a$ is a latent variable. The generated response $c$ is expected to correlate with the core information related to this \textit{anchor}.

To analyze spurious biases in MLLMs,  without loss of generality, we assume that each input from both modalities contains core features, spurious features, and noise features, following prior works \cite{sagawa2019distributionally, sagawa2020investigation, ming2022impact, xue2024understanding}. Specifically, we represent the vision input as $\mathbf{x} = [x_{\text{core}}, x_{\text{spu}}, x_{\text{noise}}]$ and the language input as $\mathbf{y} = [y_{\text{core}}, y_{\text{spu}}, y_{\text{noise}}]$. In both modalities, the core features $(x_{\text{core}}, y_{\text{core}})$ are essential for generating the intended response $c$. Spurious features $(x_{\text{spu}}, y_{\text{spu}})$ are non-essential to $c$ but may exhibit statistical correlations with it, while noise features $(x_{\text{noise}}, y_{\text{noise}})$ capture sample-specific details, which represent some minor independent perturbations in images or irrelevant contexts in texts.

\begin{figure*}[t]
    \centering
    \includegraphics[width=\textwidth]{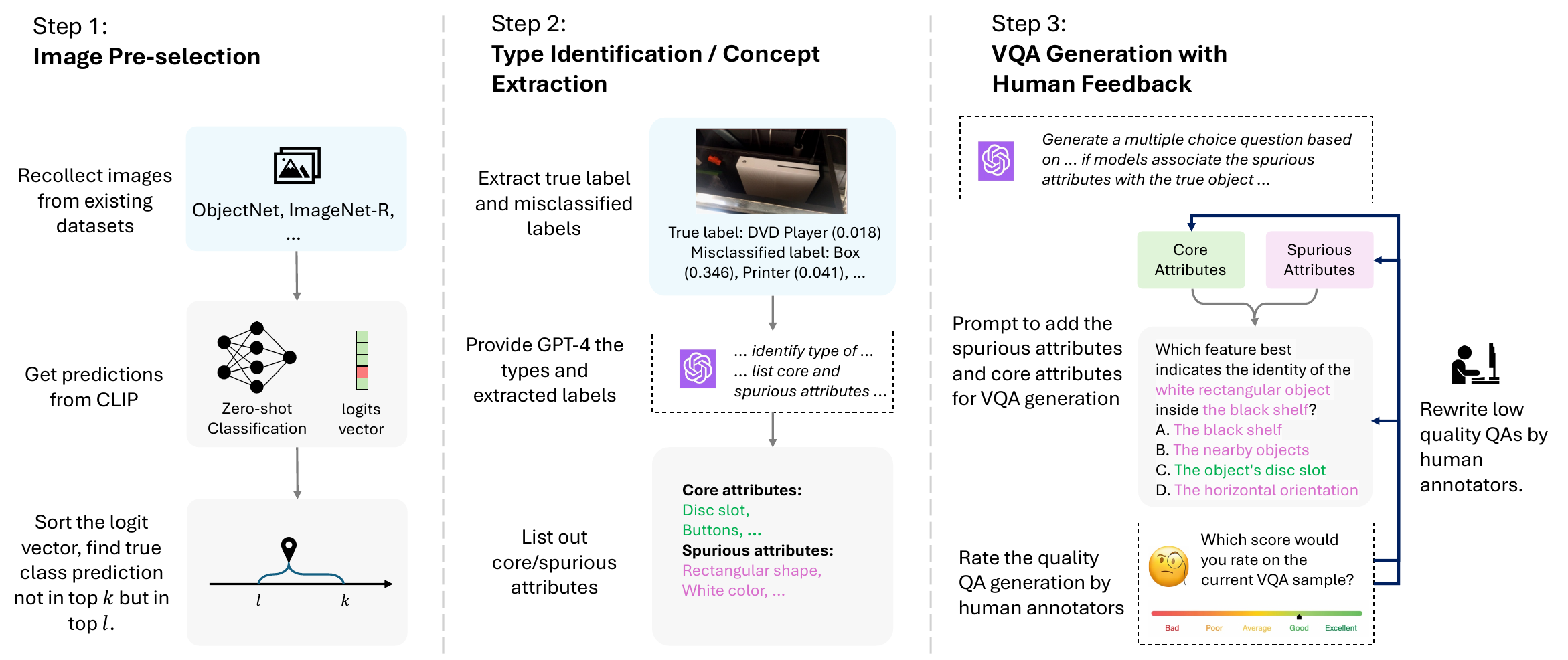}
    \caption{Construction of the \textsc{MM-SpuBench}. \textbf{Left:} Pre-select images where CLIP's true class prediction is not in the top $k$ but is in the top $l$. \textbf{Middle:} Identify spurious correlations and list core/spurious attributes. \textbf{Right:} Generate multiple-choice questions based on the spurious bias type and core/spurious attributes with human supervision.} \label{fig:bench_pipe}
\end{figure*}

\subsection{From Single Modality to Multi-modality}


We start with the analysis of single-modality spurious bias. Without loss of generality, we first consider the vision modality $\mathcal{X}$. Assume there exists a data pair $(\mathbf{x},c)$, with $\mathbf{x}$ as the vision input and $c$ as the true output, derived from a training dataset, and $\mathbf{z}$ is a spurious attribute 
of $\mathbf{x}$. Given that $\mathbf{z}$ and $c$ form a spurious correlation in the training set, the condition for a model to develop spurious bias is as follows:
$
p_{\text{train}}(\mathbf{z} | c, x_{\text{core}}) \gg p_{\text{train}}(\mathbf{z} | x_{\text{core}})
$ \cite{yang2023change}, which means that the probability of $\mathbf{z}$ given $c$ and the core feature $x_{\text{core}}$ is much higher than the probability of $\mathbf{z}$ given only the core feature $x_{\text{core}}$. Similarly, in the language modality, given a data pair $(\mathbf{y},c)$ and the associated spurious attribute $\mathbf{z}$, the condition for developing spurious bias in the language modality can be expressed as:
$
p_{\text{train}}(\mathbf{z} | c, y_{\text{core}}) \gg p_{\text{train}}(\mathbf{z} | y_{\text{core}}).
$
We extend our analysis to the multimodal setting and define multimodal spurious bias as follows.



\begin{definition}[Multimodal Spurious Bias]
\label{def:mm_spurious}
Given an input image $\mathbf{x}=[x_{\text{core}}, x_{\text{spu}}, x_{\text{noise}}]$, a text input $\mathbf{y} = [y_{\text{core}}, y_{\text{spu}}, y_{\text{noise}}]$, the desired response $c$ to the joint inputs $\mathbf{x}$ and $\mathbf{y}$, and a spurious attribute $\mathbf{z}$ shared by $x_{spu}$ and $y_{spu}$, the spurious correlations in the multimodal setting can be expressed as follows:
\begin{align}
    p(\mathbf{z}|x_{\text{core}},y_{\text{core}},c) \gg p(\mathbf{z}|x_{\text{core}},y_{\text{core}}).
    \label{eq:multimodal}
\end{align}
\end{definition}
Under this condition, an MLLM tends to develop spurious multimodal bias, which is the tendency to use the spurious correlations between spurious attributes $\mathbf{z}$ and the desired responses $c$ to generate responses given the core features in both modalities. Detailed discussions on the definition are provided in the Appendix.

\subsection{How to Reveal Multimodal Spurious Bias}
In principle, revealing spurious biases in models requires the construction of a test set in which the spurious correlations are distributionally shifted relative to those in the training data. For example, in the vision modality, a common approach \cite{sagawa2019distributionally} is to curate a test set where the spurious correlation between a spurious attribute $\mathbf{z}$ and a target $c$ becomes
$
p_{\text{test}}(\mathbf{z} | c, x_{\text{core}}) = p_{\text{test}}(\mathbf{z} | x_{\text{core}})
$, compared to $
p_{\text{train}}(\mathbf{z} | c, x_{\text{core}}) \gg p_{\text{train}}(\mathbf{z} | x_{\text{core}})
$ in the training set. Thus, the correlation between $\mathbf{z}$ and \( c \) only holds in the training data and no longer holds in the test data. Motivated by this, we propose the following multimodal data generation method to reveal multimodal spurious bias.

We first select images with the spurious attribute $\mathbf{z}$ following $p_{\text{test}}(\mathbf{z} | c, x_{\text{core}}) \approx p_{\text{test}}(\mathbf{z} | x_{\text{core}})$ in the vision modality. This can be approximated by collecting misclassified samples whose spurious attributes are unlikely to correlate with certain labels, as seen in the training data. Given the anchor (true label),  we derive the core and spurious attributes from the misclassified samples. Based on the derived attributes, we create individual VQA tasks that contain various spurious textual attributes. This process simulates $p_{\text{test}}(\mathbf{z} | c, y_{\text{core}}) \approx p_{\text{test}}(\mathbf{z} | y_{\text{core}})$. In this way, the spurious correlation learned by the model between $c$ and $\mathbf{z}$ is broken, thus causing failure if the models rely on spurious correlations for prediction. 

However, constructing such a test set requires knowing $\mathbf{z}$ \textit{a priori}, which is challenging in the multimodal scenario where only core objects are annotated in previous datasets. We address this challenge by proposing a semi-automatic curation method and producing a comprehensive human-verified VQA benchmark. We explain our curation pipeline with details in Section \ref{sec:construct}.

\section{\textsc{MM-SpuBench}: The Benchmark for Multimodal Spurious Biases}

\subsection{Types of Spurious Correlations}
We first define nine types of spurious correlations to comprehensively cover the spurious correlations in real-world data, as shown in Fig.~\ref{fig:illustration}. Although some research~\cite{gavrikov2024vision} exists with similar definitions, such as shape bias and texture bias, we are interested in diverse types of spurious correlations between attributes and core objects in images rather than focusing on a single type of bias.

\subsection{Construction of \textbf{\textsc{MM-SpuBench}}}
\label{sec:construct}
In this section, we demonstrate the three steps for the construction of \textsc{MM-SpuBench} as shown in Fig. \ref{fig:bench_pipe}.

\textit{Image pre-selection.} As shown in Fig.~\ref{fig:bench_pipe}~(left), we pre-select images with their class labels from various image classification datasets to ensure the diversity of our benchmark. ObjectNet~\cite{objectnet} serves as our primary image source due to its numerous observable spurious biases. To supplement this dataset, we collect data from descendants of ImageNet-Hard~\cite{taesiri2023zoom} that focus on domain generalization evaluation, including ImageNet-R (rendition)~\cite{hendrycks2021many}, ImageNet-Sketch~\cite{wang2019learning}, ImageNet-A~\cite{hendrycks2021natural}, and ImageNet-C~\cite{hendrycks2019benchmarking}. They complement categories of spurious biases not present in ObjectNet, such as texture/noise and relative size. We choose existing datasets rather than using parametric image generation techniques~\cite{han2024instinctive} to ensure our benchmark reflects realistic spurious biases found in the real world, avoiding additional biases that could render the benchmark results unrepresentative. See Appendix for the dataset details.

To select image data with potential spurious attributes, we use the most commonly employed vision encoder in current open-source MLLMs, \texttt{CLIP-ViT-L/14@336px}~\cite{clip}, for zero-shot classification. As evidenced by~\cite{yang2023mitigating,tong2024eyes}, CLIP models significantly suffer from the spurious correlations learned during pretraining. We utilize the logit vectors from the classification output to find samples where CLIP's true class prediction is in the range of top-$k$ and top-$l$, where $k$ and $l$ are hyperparameters to control whether the misclassification is due to spurious biases rather than annotation noise or insufficient visual cues. For each image, we record the ground truth class and top misclassified classes. Some pairs of ground truth labels and misclassified labels indicate the spurious correlations the vision encoder relies on during the training process, guiding the following design of our benchmark. For image pre-selection, we deploy $k=3, l=20$ for ObjectNet and $k=3, l=40$ for ImageNet-Hard. With this selection strategy, we curate a dataset with a total of $10,773$ image samples with labels. To retrieve a high-quality VQA subset, we deploy $k=5, l=10$ for ObjectNet and $k=3, l=40$ for ImageNet-Hard with a total of $2,400$ image/label samples.

\textit{Type identification and attribute extraction.}
In Fig.~\ref{fig:bench_pipe}~(middle), we leverage images along with their corresponding ground truth and misclassified labels to identify the types of spurious biases and understand their underlying causes. To achieve this, we employ a strong LLM (i.e., GPT-4o) as a concept generator, utilizing the chain-of-thought strategy to extract detailed and useful concept-based information from both the ground truth and misclassified labels. For each image, we generate two types of attributes: core and spurious attributes. Core attributes are generated based on the ground truth label. They describe the intrinsic properties of the core object within the image, such as shape, color, and specific distinguishable features inherent to the object. Spurious attributes are generated based on the misclassified labels. These attributes do not have direct correlations with the primary object but still influence the model's inference process, leading to spurious biases. To maintain a balanced and fair evaluation in our VQA benchmark, we limit the number of both core and spurious attributes to 5 per image, ensuring consistent evaluation and fair comparison across the dataset. Then we used the derived attributes together with the image to let the LLM determine the types of spurious biases (at most $2$) in the image. We present the type distribution in our dataset in Table~\ref{tab:stat}.

\begin{table}[!t]
\small
    \caption{Main statistics of \textsc{MM-SpuBench}. Each image may contain at most two types of spurious correlations.}
    \centering
    \resizebox{0.75\linewidth}{!}{
    \begin{tabular}{l|c}
        \hline
        Statistic & Value \\
        \hline
        Total instances & 2,400 \\ \hline
        Background (\texttt{BG}) & 1,908 (79.50\%) \\
        Texture and Noise (\texttt{TN}) & 288 (12.00\%) \\
        Co-occurring Objects (\texttt{CO}) & 903 (37.62\%) \\
        Relative Size (\texttt{RS}) & 75 (3.12\%) \\
        Colorization (\texttt{Col.}) & 248 (10.33\%) \\
        Orientation (\texttt{Ori.}) & 764 (31.83\%) \\
        Lighting and Shadows (\texttt{LS}) & 103 (4.29\%) \\
        Perspective and Angle (\texttt{PA}) & 100 (4.17\%) \\
        Shape (\texttt{Sha.}) & 401 (16.71\%) \\
        \hline
    \end{tabular}
    }
    \label{tab:stat}
\end{table}

\textit{Visual Question Answering (VQA) curation.}
In Fig.~\ref{fig:bench_pipe}~(right), we build upon the identified core and spurious attributes to create VQA pairs that evaluate a model's robustness to multimodal spurious biases. To reduce crowdsourcing, we adopt a semiautomatic improvement process for VQA curation. Using selected images and their core/spurious attributes, we design prompts that incorporate spurious attributes into the questions and use core attributes to generate one correct option referring to the anchor. A strong LLM uses this information to produce multiple-choice questions to test whether a model can identify the golden response in the presence of spurious attributes. These questions avoid direct cues to the core attributes or true labels, instead describing the core object with its spurious attributes and spatial position. Each question randomly incorporates the derived core and spurious attributes from the previous step, with only one ground truth answer and three misleading options. Following the data generation, five human annotators review and rate QA triplets on a scale of 1 to 5. Based on this feedback, annotators revise the core/spurious attributes, QA pairs, and ground truth answers with human-grounded knowledge. An overview of the \textsc{MM-SpuBench} data is illustrated in Fig. \ref{fig:illustration}.

\subsection{Beyond VQA: Conditional Generation Likelihood}

The standard VQA accuracies on our dataset provide high-level insight into the robustness of the MLLMs against spurious biases. To benchmark MLLMs at a finer level, with less dependency on the MLLMs' overall instruction-following and visual recognition capabilities, we calculate the MLLMs' preferences for all triplets of desired responses, textual attributes, and visual inputs. Inspired by \citet{mcmilin2022selection} on using single-token distribution to investigate correlations in Pretrained Bidirectional Language Models, we adopt the generation probability as a lens to reveal the learned correlations of MLLMs. 


We extend the single-token generation probability to MLLMs and adapt for causal generation as follows: with MLLMs that consider visual representations as a pretext to the generation, the probability of the next token is $P(s_i | \textbf{s}_{i-1}, \textbf{v})$, where $s_i$ is the current generated token, $\textbf{s}_{i-1}$ is all text tokens before position $i$ and $\textbf{v}$ is a vector of visual features. For a fixed textual sequence $\textbf{o}:= o_0,o_1,\ldots,o_k$, we compute the logarithmic likelihood of text generation in the model from a pretext pair $(\textbf{s}_{i-1}, \textbf{v})$: 
\begin{align}
  \text{MLLM}(\textbf{o} | \textbf{s}_{i-1}, \textbf{v}) &= \prod_{i=l}^{l+k}  P(s_i = o_i | \textbf{s}_{i-1}, \textbf{v})  
\end{align}

At inference time, we formulate a generation prompt, consisting of a system prompt, a user prompt, and an assistant prefix, as input for the generation of MLLM. Denote $g(\cdot)$ as the prompt function that takes an input attribute and produces a natural text sequence containing the input attribute. The Conditional Generation Likelihood under an input attribute text $\textbf{t}$ of the caption of the ground truth label as the true response $\textbf{c}$ is 
\begin{align}
  \text{CGL}(\textbf{c}|\textbf{t}, \textbf{v}) &= \log[\text{MLLM}(\textbf{c}|g(\textbf{t}), \textbf{v})] 
  \label{eq:cgl}
\end{align}

We propose Conditional Generation Likelihood Advantage (CGLA) as a metric to detect the learned spurious bias of an MLLM. We calculate the Conditional Generation Likelihood following Equation \ref{eq:cgl}. Then, CGLA of the true response $\textbf{c}$ of one attribute text $\textbf{t}_1$ over another attribute text $\textbf{t}_2$ is calculated by subtracting CGL$(\textbf{c},\textbf{t}_2)$ from CGL$(\textbf{c},\textbf{t}_1)$, which is defined as follows:
\begin{align}
    \text{CGLA}(\textbf{c}| \textbf{t}_1; \textbf{t}_2, \textbf{v}) = \text{CGL}(\textbf{c}| \textbf{t}_1, \textbf{v}) - \text{CGL}(\textbf{c}|\textbf{t}_2, \textbf{v})
    \label{eq:cgla_one}
\end{align}

 In our experiments, given an annotated set of spurious and core attributes, denoting $\textbf{t}_{\text{core}}$ as the core attribute, $\textbf{t}_{\text{spu}}$ as a spurious attribute and $\mathcal{T}_{\text{spu}}$ as the set of spurious attributes, we consider the CGLA difference between the core attribute and the spurious attribute that makes generating the object class most probable as a measurement of the model's preference toward spurious attributes as a condition for the corresponding object class.
We specifically select the spurious attribute that maximizes the generation probability because it represents the most tempting shortcut available to the model, allowing our metric to quantify the model's bias under a worst-case condition, providing a stringent test of its robustness. We formally define it as follows:
\begin{align}
    \text{CGLA}_{\min}(\textbf{c} | \textbf{t}_{\text{core}}; \mathcal{T}_{\text{spu}}, \textbf{v}) &= \min_{\textbf{t}_{\text{spu}} \in \mathcal{T}_{\text{spu}}} \left( \text{CGLA}(\textbf{c} |\textbf{t}_{\text{core}}; \textbf{t}_{\text{spu}}, \textbf{v}) \right)
  \label{eq:cgla}
\end{align}

Based on the causal language modeling objective of MLLMs, CGLA measures the likelihood difference between attributes as conditions for generating a particular text sequence as output. Under our evaluation setup, a positive CGLA$_{\min}$ means that the core attribute is the most competitive condition for generating the object label among the spurious attributes, and a negative CGLA$_{\min}$ means that at least one spurious attribute is a better condition for generating the object label. Therefore, we apply the unit step function to CGLA$_{\min}$. When averaged over the dataset, this metric measures an MLLM's accuracy in modeling the correct condition-anchor relationship under our generative evaluation setting. Given the unit step function $\mathds{1}(\cdot)$, the generative accuracy CGLA$_{\text{acc}}$ is defined:
\begin{align}
   \text{CGLA}_{\text{acc}}(\textbf{c} | \textbf{t}_{\text{core}}; \mathcal{T}_{\text{spu}}, \textbf{v}) &= \mathds{1} \left(\text{CGLA}_{\min}(\textbf{c} | \textbf{t}_{\text{core}}; \mathcal{T}_{\text{spu}}, \textbf{v}) \right)
   \label{eq:cgla_acc}
\end{align}

\section{Results and Analysis}

\begin{table*}[t]
\begin{center}

\newcommand*{\SetGradientLimits}[2]{%
            \renewcommand*{\MinNumber}{#1}%
            \renewcommand*{\MaxNumber}{#2}%
        }

\caption{Zero-shot results of MLLMs on \textsc{MM-SpuBench}. All numbers are percentage accuracies. {\setlength{\fboxsep}{1pt}\colorbox{goodgreen!100}{Green}} color indicates higher accuracy, and {\setlength{\fboxsep}{1pt}\colorbox{goodred!100}{red}} color indicates lower accuracy. The dashes indicate that the proprietary models' backbones are not applicable.} 
\label{tab:open_model_acc}
\begin{tabularx}{\linewidth}{lc H{0}{100}H{0}{100}H{0}{100}H{0}{100}H{0}{100}H{0}{100}H{0}{100}H{0}{100}H{0}{100}H{0}{100}}
\toprule
\multicolumn{1}{c}{\multirow{2}{*}{MLLM}} &
  \multicolumn{1}{c}{\multirow{2}{*}{LLM Backbone}} &
  \multicolumn{9}{c}{\textsc{MM-SpuBench}} &
  \multicolumn{1}{c}{\multirow{2}{*}{Average}} \\ \cmidrule{3-11}
 &
   &
  \multicolumn{1}{c}{\texttt{BG}} &
  \multicolumn{1}{c}{\texttt{TN}} &
  \multicolumn{1}{c}{\texttt{CO}} &
  \multicolumn{1}{c}{\texttt{RS}} &
  \multicolumn{1}{c}{\texttt{Col.}} &
  \multicolumn{1}{c}{\texttt{Ori.}} &
  \multicolumn{1}{c}{\texttt{LS}} &
  \multicolumn{1}{c}{\texttt{PA}} &
  \multicolumn{1}{c}{\texttt{Sha.}} &
  \multicolumn{1}{c}{}
   \\ \midrule\midrule
   Gemini 1.5 Pro \cite{team2023gemini} & ---  & 60.12 & 55.35 & 63.46 & 50.28 & 53.25 & 62.86 & 60.38 & 48.15 & 54.79 & 58.06 \\
   Claude 3 Haiku \cite{claude3_family} & --- & 55.45 & 53.77 & 57.12 & 40.22 & 45.12 & 55.71 & 47.17 & 37.04 & 39.85 & 52.06 \\
   Claude 3  Sonnet \cite{claude3_family} & --- & 78.06 & 76.57 & 81.35 & 61.45 & 65.85 & 81.43 & 75.47 & 59.26 & 60.92 & 74.82 \\
   Claude 3 Opus \cite{claude3_family} & --- & 80.43 & 76.10 & 83.65 & 64.80 & 66.67 & 82.86 & 83.02 & 70.37 & 67.82 & 77.18 \\
   GPT-4V \cite{achiam2023gpt} & --- & 83.58 & 82.39 & 85.33 & 67.60 & 72.65 & 81.43 & 84.91 & 70.37 & 73.36 & 80.90 \\
   GPT-4o \cite{achiam2023gpt} & --- & 80.64 & 81.13 & 83.85 & 60.89 & 69.39 & 80.00 & 83.02 & 65.43 & 67.18 & 77.97 \\ 
   
   \midrule
   
    InternVL3-8B~\cite{zhu2025internvl3exploringadvancedtraining} & Qwen2.5-7B~\cite{Qwen2.5-VL}
    & 74.48 & 70.87 & 78.01 & 58.87 & 60.42 & 73.79 & 70.67 & 68.00 & 67.33 & 71.92 \\
    InternVL3-14B~\cite{zhu2025internvl3exploringadvancedtraining}  &Qwen2.5-14B~\cite{Qwen2.5-VL}
    & 81.55 & 80.51 & 82.85 & 64.92 & 66.32 & 75.73 & 80.00 & 69.00 & 64.09 & 77.92 \\
     InternVL3-38B~\cite{zhu2025internvl3exploringadvancedtraining} & Qwen2.5-32B~\cite{Qwen2.5-VL}
    & 86.11 & 83.28 & 88.22 & 73.79 & 70.14 & 85.44 & 81.33 & 76.00 & 74.56 & 83.04 \\ 
    Llama-3.2-VI-11B~\cite{meta2024llama32} &Llama-3.1-8B~\cite{meta2024llama31}
    & 74.58 & 73.53 & 76.57 & 59.27 & 62.85 & 68.93 & 74.67 & 63.00 & 61.35 & 71.71 \\
    LLaVA-v1.5-7B~\cite{liu2024visual} & Llama-2-7B~\cite{touvron_llama_2023} 
    & 41.19 & 41.31 & 39.53 & 33.47 & 34.38 & 40.78 & 38.67 & 38.00 & 35.16 & 39.54 \\
    LLaVA-v1.5-13B~\cite{liu2024visual} & Llama-2-13B~\cite{touvron_llama_2023} 
    & 63.31 & 62.57 & 63.48 & 39.52 & 46.18 & 55.34 & 56.00 & 50.00 & 47.63 & 59.08 \\ 
    LLaVA-v1.6-mis-7B~\cite{liu2024visual} & Mistral-7B~\cite{jiang_mistral_2023} 
    & 39.83 & 39.20 & 40.71 & 32.26 & 38.54 & 38.83 & 41.33 & 42.00 & 33.67 & 38.88 \\
    LLaVA-v1.6-vic-13B~\cite{liu2024visual} & Vicuna-13B~\cite{zheng_judging_2023} 
    & 24.53 & 26.25 & 24.74 & 20.97 & 23.96 & 28.16 & 24.00 & 19.00 & 23.44 & 24.58 \\
    LLaVA-v1.6-34B~\cite{liu2024visual} & Hermes-Yi-34B~\cite{ai_yi_2024} 
    & 75.79 & 72.43 & 78.14 & 58.87 & 60.42 & 67.96 & 70.67 & 64.00 & 63.09 & 72.17 \\
    Qwen2-VL-7B~\cite{bai2023qwen} & QwenLM-7B~\cite{bai_qwen_2023} 
    & 73.95 & 69.44 & 74.21 & 52.02 & 56.25 & 69.90 & 73.33 & 61.00 & 55.36 & 69.04\\
    Qwen2.5-VL-7B~\cite{Qwen2.5-VL} & Qwen2.5-7B~\cite{Qwen2.5-VL}
    & 71.59 & 67.55 & 73.43 & 53.63 & 58.33 & 73.79 & 74.67 & 56.00 & 61.85 & 68.38\\
    \bottomrule
\end{tabularx}
\end{center}
\end{table*}

\begin{table}[t]

\caption{Average CGLA$_{\min}$ and CGLA$_{\textnormal{acc}}$ of open-source MLLMs on attributes derived from \textsc{MM-SpuBench}. Higher values in {\setlength{\fboxsep}{1pt}\colorbox{goodgreen!100}{green}} indicate well-aligned attribute-anchor relations. Lower values in {\setlength{\fboxsep}{1pt}\colorbox{goodred!100}{red}} indicate misalignment.} 

\label{tab:open_model_cgla}
\begin{center}
\newcommand*{\SetGradientLimits}[2]{%
            \renewcommand*{\MinNumber}{#1}%
            \renewcommand*{\MaxNumber}{#2}%
        }

\begin{tabularx}{\linewidth}{lH{-3.3}{3.3}H{-3.3}{3.3}H{0}{100}H{0}{100}}
\toprule
\multirow{2}{*}{MLLM} &
  \multicolumn{2}{c}{CGLA$_{\min}$} &
  \multicolumn{2}{c}{CGLA$_{\text{acc}}$ (\%)}  
 \\ \cmidrule{2-5}
 &
  \multicolumn{1}{c}{User} &
  \multicolumn{1}{c}{Asst.} &
  \multicolumn{1}{c}{User} &
  \multicolumn{1}{c}{Asst.}
   \\ 
   
   \midrule
   \midrule
   
    InternVL3-8B~\cite{zhu2025internvl3exploringadvancedtraining} 
    & 1.01 & 2.95 & 61.62& 78.21 \\
     InternVL3-14B~\cite{zhu2025internvl3exploringadvancedtraining} 
    & 0.52 & 1.65  & 61.50& 69.54\\
     InternVL3-38B~\cite{zhu2025internvl3exploringadvancedtraining} 
    &  1.64 & 3.27 & 66.62& 79.79\\
    Llama-3.2-VI-11B~\cite{meta2024llama32}
    & 1.67 & 2.40 & 63.75 & 71.29\\
    LLaVA-v1.5-7B~\cite{liu2024visual} 
    & 0.06 & -0.06 & 60.38 & 38.46\\
    LLaVA-v1.5-13B~\cite{liu2024visual} 
    & 0.21 &  1.14 & 57.83 & 68.62\\
    LLaVA-v1.6-mis-7B~\cite{liu2024visual} 
    & 1.41 & 2.84 & 71.12 & 80.96\\
    LLaVA-v1.6-vic-13B~\cite{liu2024visual} 
    & 0.13 & 0.08 & 53.04 & 44.17\\
    LLaVA-v1.6-34B~\cite{liu2024visual} 
    & 0.16 & 1.68 & 52.92& 71.29\\
    
    \bottomrule
\end{tabularx}
\end{center}
\end{table}
\subsection{Baselines}

For proprietary MLLMs, we chose Gemini 1.5 Pro~\cite{team2023gemini}, GPT-4V/GPT-4o~\cite{achiam2023gpt}, and the Claude 3 family models (Haiku, Sonnet, Opus)\cite{claude3_family}, which are the mainstream MLLMs. The input for these models consists of a system prompt and a format prompt that describes the task and the question with four options, while the expected output includes the predicted option and an explanation to help us understand the reasoning processes.
For open-source MLLMs, following previous works~\cite{liu2023mmbench,tong2024eyes}, we select current state-of-the-art models that excel in general VQA tasks, including InternVL3~\cite{zhu2025internvl3exploringadvancedtraining}, Llama3.2~\cite{meta2024llama32}, LLaVA~\cite{liu2024visual}, and Qwen-VL~\cite{bai2023qwen, Qwen2.5-VL}, with variants of LLM backbones. The input for these models includes a system prompt that describes the task and the question with four options, with the expected output being only the option following a vanilla multiple-choice VQA setup. 



\subsection{Experimental Details}
\label{sec:imple_detail}

\paragraph{Standard VQA Experiment.} To ensure fair comparisons, we shuffle the answer choices for each question to eliminate option biases across MLLM models. For open-source models, all inference experiments are run on NVIDIA A100 and A6000 GPUs. Each experiment is repeated with three different random seeds, and the reported values are the average of these runs. For consistency and reproducibility, we perform \textit{greedy sampling}. Due to variations in the capabilities of each model, we design separate prompts to ensure the models can output the correct format of choice from our benchmark. To assess the MLLMs' performances on \textsc{MM-SpuBench}, we use accuracy as the metric to determine MLLMs' robustness to spurious biases. For proprietary models, we only include cases when the model output follows the choice format. Among open-source models, Llama3.2~\cite{meta2024llama32} and LLaVA-v1.6~\cite{liu2024visual} models frequently fail to follow the instructions and produce text descriptions of their answers. To compensate for this, we concatenate the generation with ``\textit{Choice:}'' and generate the next token only among the choice letters as their final answer. Detailed VQA prompts are included in the Appendix.

\paragraph{Generative Likelihood Experiment.} From each VQA triplet of image, question, and choices, we obtain $3$ spurious attributes and $1$ core attribute, and perform generative evaluations on the quartet of image, object class, core attribute, and set of spurious attributes. We formulate two chat templates, the User template and the Assistant template. The User template constructs a natural single-round open-ended visual chat task, where the user deliberately includes one attribute in the text input and asks about the type of the anchor object. The Assistant template follows a similar single-round open-ended visual chat task, where the user asks for observation in the image and the classification of the anchor object. In this template, the prefix of assistant generation is fixed to texts that establish a relationship between one attribute and the classification. The detailed templates are included in the Appendix. We use these two templates as the prompting function in Equation~\ref{eq:cgl} and calculate the CGL of the given object label text under each attribute. The CGLA$_{\min}$ is calculated following Equation~\ref{eq:cgla} and CGLA$_{\text{acc}}$ is calculated following Equation~\ref{eq:cgla_acc}. The reported statistics are averaged over the dataset. 

\subsection{Main Results}
\paragraph{Overall VQA Accuracy on \textsc{MM-SpuBench}.} Based on the results in Table~\ref{tab:open_model_acc}, we observe that MLLMs exhibit varying degrees of spurious bias. The most recent open-sourced model, InternVL3~\cite{zhu2025internvl3exploringadvancedtraining}, performs comparably to the best closed-sourced model tested. Across different types of spurious bias, we found significant variations in the MLLMs' ability to address each type. They perform better in the \texttt{BG} and \texttt{CO} types, while their performance is notably subpar in the \texttt{RS} and \texttt{Col.} types. This gap indicates that when prompted to choose the exact attribute that distinguishes a class of objects, MLLMs tend to fail more frequently when certain types of spurious attributes are present. Assuming the attribute selection of the MLLMs aligns with its implicit emergent reasoning process, we argue that certain attribute types are more tempting for the MLLMs, indicating the potential presence of such spurious biases in the MLLMs.

\textit{Generative Likelihood Analysis.} 
The average conditional generation statistics are reported in Table~\ref{tab:open_model_cgla}, column text ``User'' and ``Asst.'' indicates the metric is computed with the \textit{User prompt} template and the \textit{Assistant prompt} template, respectively. We remove Qwen~\cite{bai2023qwen, Qwen2.5-VL} models from our experiments because they do not produce logits within the floating point range, making calculation impossible. In most models, the \textit{Assistant prompt} template produced better alignment between the attributes and the anchor concept. We attribute this advantage to the more detailed and deliberate causal relationship in the \textit{Assistant prompt}. Such a relationship is not applicable in the \textit{User prompt} as the assistant's prefix does not contain the attribute text. We notice that LLaVA-v1.5-7B~\cite{zhang2025llava} and LLaVA-v1.6-vic-13B~\cite{zhang2025llava} do not benefit from this stronger conditioning, which can be caused by possible failure in instruction-following or disturbance caused by the long custom assistant prefix. The \textit{User prompt} provides a more natural evaluation, as the user can input arbitrary text in user-MLLM interactions. Based on the \textit{User prompt} results, most models demonstrated, on average, a positive core condition advantage, with InternVL3-38B~\cite{zhu2025internvl3exploringadvancedtraining}, Llama-3.2-VI-11B~\cite{meta2024llama32}, and LLaVA-v1.6-mis-7B~\cite{zhang2025llava} performing the best. On per-instance confidence, these models show high CGLA$_{\min}$ values greater than $\log(4)\approx1.38$, indicating that on average these MLLMs generate the anchor text $4$ times more likely when presented with a core attribute than with a spurious attribute. Focusing on preference alignment, these models demonstrated the highest CGLA$_{\text{acc}}$ among tested models. 

\textit{Combined Analysis.}
The overall performances of the MLLMs under the VQA test and the generative test are overall aligned, with the exceptions of LLaVA-v1.6-mis-7B~\cite{zhang2025llava} and InternVL-14B~\cite{zhu2025internvl3exploringadvancedtraining}. We see LLaVA-v1.6-mis-7B~\cite{zhang2025llava} performs well on generative test but produced low VQA accuracy. A slight difference between VQA and Generative tests may cause this discrepancy: the VQA test does not show the object anchor, whereas in the generative test, though the model is not exposed to the anchor when processing the attribute, the anchor information is given in the test pipeline. Therefore, the Generative test is an easier task, since the MLLMs do not need to resolve the true object class. Our argument is supported by higher accuracies across several models in Table~\ref{tab:open_model_cgla}. However, InternVL3-14B~\cite{zhu2025internvl3exploringadvancedtraining} is an exception. Its degraded accuracy from VQA to Generative test may indicate a misalignment of attributes and anchors that is not uncovered by high-level VQA testing, stressing the importance of generative evaluations.

\subsection{Visualization of Spurious Bias}
To illustrate the influence of spurious features on MLLM reasoning and validate \textsc{MM-SpuBench}’s ability to reveal spurious biases learned by the models, we visualize Grad-CAM heatmaps~\cite{selvaraju2020grad} and token-level text attention heatmaps of LLaVA-v1.5-7B~\cite{zhang2025llava} and LLaVA-v1.5-13B~\cite{zhang2025llava} on examples in \textsc{MM-SpuBench}. We show one case in Fig.~\ref{fig:gradcam_main} and more visualizations in the Appendix.
This example demonstrates that spurious attributes mislead the model. The Grad-CAM heatmaps pay more attention to irrelevant elements in the bathroom scene, such as the ``toothpaste'' and other background objects. Although the core feature of the object (the gripping structure) is present in the image, the model’s focus is dispersed across spurious features. The token-level text attention heatmap further supports this observation, where generated text reasoning heavily emphasizes ``bathroom'' and other contextual elements. This agreement between visual and textual attention confirms that the model contains some spurious biases, which hinder it from learning the core attributes of the object and lead to incorrect predictions.
Additionally, we show that increasing the model size from 7B to 13B can reduce the influence of spurious features to some extent. The 13B model reduces the focus on spurious attributes and focuses more on the core attribute ``pair of tweezers''. However, we can still observe the focus of spurious attributes. It indicates that scaling alone is not sufficient to mitigate the impact of spurious correlations. This underscores the importance of our benchmark in evaluating and improving model robustness in complex multimodal scenarios.

\begin{figure}[t]
    \centering
    \includegraphics[trim=498px 10px 523px 1px, clip, width=0.9\linewidth]{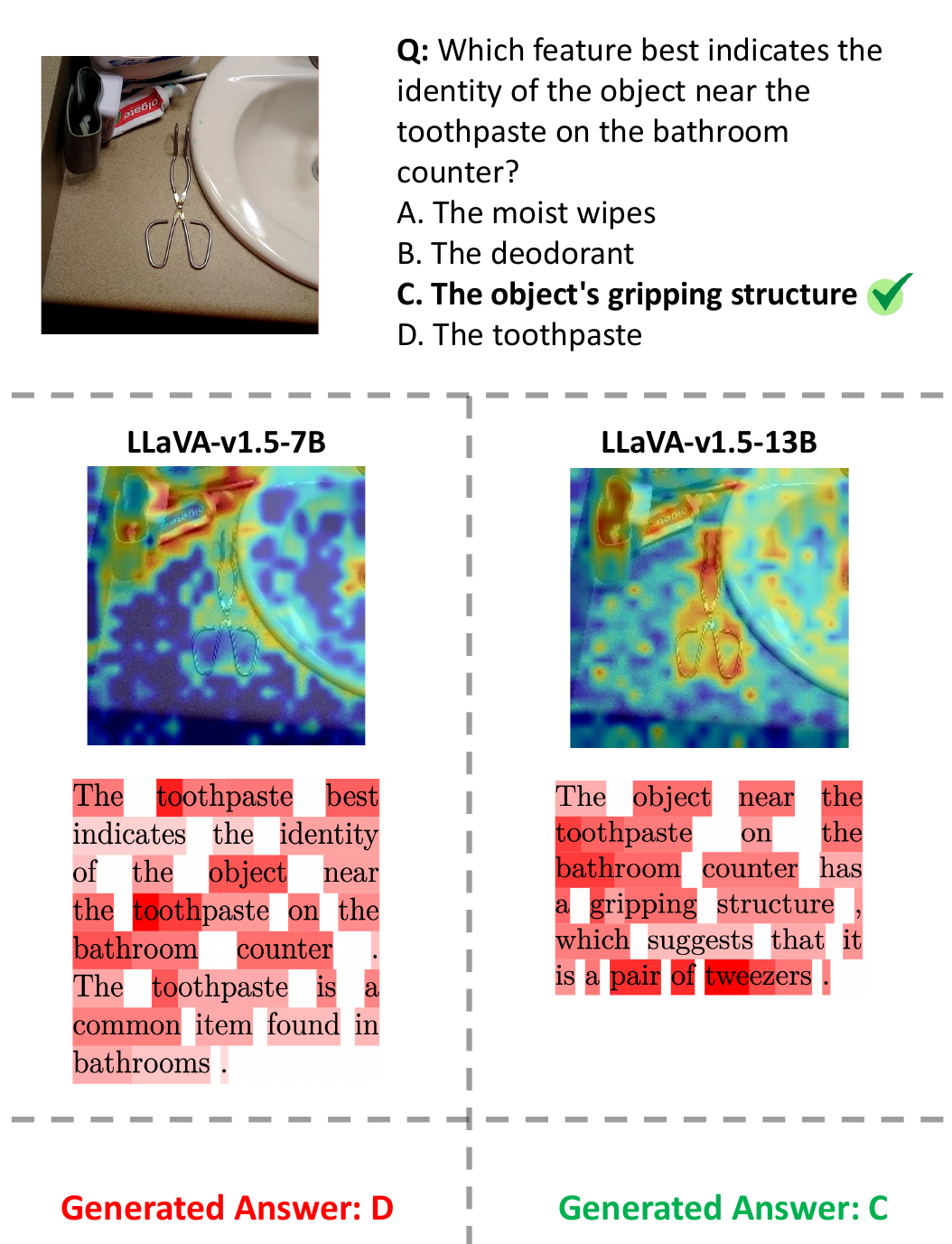} 
    \caption{GradCAM Visualization and Text Attention Heatmap of \textsc{MM-SpuBench} on LLaVA 7B and 13B. The \textit{anchor} objects in the image is: \texttt{tweezers}.}
    \label{fig:gradcam_main}
\end{figure}

\section{On Mitigating Spurious Correlation Effects}
With \textsc{MM-SpuBench}, we explore the simple strategy of textual prompting as a preliminary investigation into the MLLMs' implicit and explicit reasoning abilities as a key to mitigate spurious biases in multimodal question answering. 

Following \citet{lu2024seeing} and \citet{li2025core}, we explore simple prompting and evaluate its effectiveness in improving the performance of MLLMs on \textsc{MM-SpuBench}. We design and test three prompting strategies, namely: guiding, explain, and no-bias. The \textit{guiding prompt} aims to break down the reasoning process and ask the MLLM to follow it. The \textit{explain prompt} directly asks the model to give reasons first and then provide an answer. The \textit{no-bias prompt} reminds the MLLM not to fall for unrelated and spurious cues when answering. We show the prompt details in the Appendix Table \ref{tab:prompting_strategy}. The idea is to implicitly or explicitly exploit the MLLMs' reasoning capacity to identify attributes that do not constitute the anchor's core features. The benchmark results for all tested model-prompt pairs are reported in Table \ref{tab:prompt_accuracy}, and per-category accuracy is reported in the Appendix. We exclude LLaVA-v1.6-34B~\cite{zhang2025llava} due to its prohibitive computational requirements for explicit reasoning.

The \textit{vanilla prompt} yielded the best result for Qwen2-VL-7B~\cite{Qwen2.5-VL}. For this model, longer self-generated contexts introduced by our prompt could distract them from the core information. In addition, lower \textit{no-bias prompt} accuracy indicates limited implicit reasoning capacity. The \textit{no-bias prompt} was effective for InternVL-3-8B~\cite{zhu2025internvl3exploringadvancedtraining}, InternVL-3-14B~\cite{zhu2025internvl3exploringadvancedtraining}, and LLaVA-1.6-vic-13B~\cite{zhang2025llava}. It works well for models with strong implicit reasoning ability that understand and follow meta-instructions about the decision-making process. The \textit{explain prompt} is shown to increase the VQA accuracy of Llama-3.2-VI-11B~\cite{meta2024llama32}. It works best for models with strong explicit reasoning ability and long-context capacity. The \textit{guiding prompt} improved the accuracy of InternVL-8B~\cite{zhu2025internvl3exploringadvancedtraining} the most, and it also improved the performance of Llama-3.2-VI-11B~\cite{meta2024llama32}. The rest of the tested models do not show significant performance gain from simple prompting, motivating research in advanced mitigation methods. 

\begin{table}[!ht]
\caption{Overall Accuracy (\%) by Model and Prompt Type.}
\label{tab:prompt_accuracy}
\centering
\small 
\setlength{\tabcolsep}{5pt} 
\resizebox{0.95\linewidth}{!}{
\begin{tabular}{lcccc}
\toprule
\textbf{Model} & \textbf{Vanilla} & \textbf{Guiding} & \textbf{No Bias} & \textbf{Explain} \\
\midrule
\midrule
InternVL3-8B~\cite{zhu2025internvl3exploringadvancedtraining} 
& 71.92 & \textbf{75.46} & 74.96 & 72.67 \\
InternVL3-14B~\cite{zhu2025internvl3exploringadvancedtraining} 
& 77.92 & 74.88 & \textbf{79.75} & 76.96 \\
InternVL3-38B~\cite{zhu2025internvl3exploringadvancedtraining} 
& 83.04 & 82.62 & \textbf{83.29} & 82.96 \\
\midrule
Llama-3.2-VI-11B~\cite{meta2024llama32} 
& 71.71 & 72.92 & 72.75 & \textbf{74.25} \\
\midrule
LLaVA-v1.5-7B~\cite{liu2024visual} 
& 39.54 & 39.62 & 38.38 & \textbf{39.67} \\
LLaVA-v1.5-13B~\cite{liu2024visual}
& 59.08 & 55.71 & 59.25 & \textbf{59.75} \\
LLaVA-v1.6-mis-7B~\cite{liu2024visual}
& 38.88 & 39.00 & \textbf{39.08} & 33.83 \\
LLaVA-v1.6-vic-13B~\cite{liu2024visual} 
& 24.58 & 25.83 & \textbf{38.04} & 35.21 \\
\midrule
Qwen2-VL-7B~\cite{bai2023qwen} 
& \textbf{69.04} & 56.71 & 67.04 & 65.46 \\
Qwen2.5-VL-7B~\cite{Qwen2.5-VL}  
& 68.38 & 60.42 & \textbf{68.79} & 61.54 \\
\bottomrule
\end{tabular}
}

\end{table}

\section{Conclusion}

In this work, we systematically investigated the prevalence and impact of spurious biases in MLLMs. We introduced \textsc{MM-SpuBench}, a comprehensive benchmark designed to evaluate the robustness of MLLMs to spurious biases. This benchmark systematically measures how well these models distinguish between core and spurious features with two metrics, providing a useful framework for understanding and quantifying spurious biases. Our findings show that current MLLMs, particularly smaller models or those relying on basic modality alignment techniques, often fail to integrate visual and language modalities in multimodal tasks effectively. The evaluation results and visualization indicate that both open-source and proprietary MLLMs still rely on spurious correlations to various degrees, highlighting the need for improved multimodal alignment techniques and more robust architectures. We hope that \textsc{MM-SpuBench} will drive further research in this field, leading to the development of more robust and reliable multimodal LLMs.

\section*{Acknowledgments}
This work is supported in part by the US National Science Foundation under grants CCF-2217071, CNS-2213700, IIS-2106913.  Any opinions, findings, and conclusions or recommendations expressed in this material are those of the
author(s) and do not necessarily reflect the views of the National Science Foundation.

\bibliographystyle{ACM-Reference-Format}
\bibliography{bib/cao, bib/ma, bib/ye, bib/liu}

\appendix

\section*{Appendix}

\begin{table*}[t]
\caption{Types of spurious correlations categorized in \textsc{MM-SpuBench}.} \label{tab:types}
\begin{center}
\adjustbox{max width=\textwidth}{
\begin{tabular}{|l|p{11cm}|}
\hline
\textbf{Type} & \textbf{Description} \\ \hline
Background (\texttt{BG}) & Occurs when the model relies on background context instead of the subject, e.g., identifying animals by natural backgrounds and failing in urban settings. \\ \hline

Texture and Noise (\texttt{TN}) & Arises when the model focuses on textures or noise patterns instead of shapes. E.g., misclassifying fruits due to changes in surface texture. \\ \hline
Co-occurring Objects (\texttt{CO}) & Happens when the model associates frequently appearing objects together. E.g., labeling any scene with a microwave as a kitchen. \\ \hline
Relative Size (\texttt{RS}) & Occurs when the model uses the relative size of objects as a cue. E.g., misclassifying a toy car as a real car due to a close-up perspective. \\ \hline
Colorization (\texttt{Col.}) & Related to reliance on specific colors for predictions. E.g., failing to recognize bananas that are green or brown. \\ \hline
Orientation (\texttt{Ori.}) & Arises when the model depends on the orientation of objects. E.g., struggling with faces not shown upright or from side profiles. \\ \hline
Lighting and Shadows (\texttt{LS}) & Occurs when predictions are influenced by lighting conditions or shadows. E.g., misclassifying objects in images with different lighting conditions. \\ \hline
Perspective and Angle (\texttt{PA}) & Emerges when the model relies on the viewing angle of objects. E.g., car recognition failing with top-down or oblique views. \\ \hline
Shape (\texttt{Sha.}) & Arises when an object has an unusual shape resembling another object. E.g., misidentifying a deformed fruit as a different type due to shape similarity. \\ \hline
\end{tabular}
}
\end{center}

\end{table*}

\section{More Information of the Data}
\label{sec:data}
\subsection{Public Availability}
The \textsc{MM-SpuBench} dataset is publicly accessible at \url{https://huggingface.co/datasets/mmbench/MM-SpuBench}. The benchmark is licensed under the MIT License; however, users should refer to the licenses of the individual data sources outlined in the following section. The benchmark will be continuously updated to incorporate new developments and feedback from the community. Benchmark construction and evaluation code are included in the supplementary material, and the repository will be made available on GitHub after the anonymous review period. 
For a more comprehensive understanding of the benchmark, additional data instances are provided in Table~\ref{tab:data_instance_part1} and Table~\ref{tab:data_instance_part2}. 
Detailed descriptions and examples of the spurious correlation types defined in the main paper are available in Table~\ref{tab:types}.

\subsection{Data Sources and Licenses}
\paragraph{ObjectNet}
ObjectNet is a vision dataset with 50,000 images, specifically designed to test object recognition systems under varied conditions. It includes 313 object classes and controls for rotation, background, and viewpoint. This dataset reveals significant performance drops, showing real-world challenges and difficulties in transfer learning. ObjectNet is free for both research and commercial use, with the following restrictions:
\begin{enumerate}
    \item ObjectNet cannot be used to tune the parameters of any model.
    \item Individual images from ObjectNet must include their 1-pixel red border when posted online.
\end{enumerate}
The license details can be found at \url{https://objectnet.dev/download.html}.

\paragraph{ImageNet} 
ImageNet is a comprehensive visual database used for visual object recognition research, containing millions of labeled images across thousands of categories. It serves as a key benchmark for evaluating computer vision algorithms and advancing deep learning research. The license details for ImageNet are available at \url{https://www.image-net.org/download.php}.

\paragraph{ImageNet-R(endition)}
ImageNet-R is a subset of ImageNet-1K classes with art, cartoons, graffiti, embroidery, graphics, origami, paintings, patterns, plastic objects, plush objects, sculptures, sketches, tattoos, toys, and video game renditions of ImageNet classes. It contains renditions of 200 ImageNet classes, with a total of 30,000 images. This dataset is available under the MIT License at \url{https://github.com/hendrycks/imagenet-r}.

\paragraph{ImageNet-A} 
ImageNet-A contains real-world, unmodified examples that cause significant performance degradation in machine learning models. The dataset is available under the MIT License at \url{https://github.com/hendrycks/natural-adv-examples}.

\paragraph{ImageNet-C}
The ImageNet-C dataset consists of 15 types of corruptions applied to ImageNet validation images, categorized into noise, blur, weather, and digital, each with five severity levels, resulting in 75 distinct corruptions. This dataset is available under the Apache License 2.0 at \url{https://github.com/hendrycks/robustness}.

\paragraph{ImageNet-Sketch}
ImageNet-Sketch includes 50,000 images, with 50 sketches for each of the 1,000 ImageNet classes. These images are gathered using Google Image searches with the query "sketch of CLASS" in black and white. The dataset is under the MIT License at \url{https://github.com/HaohanWang/ImageNet-Sketch}.

\paragraph{ImageNet-ReaL}
ImageNet-ReaL offers "Re-Assessed" (ReaL) labels with multi-label and more accurate annotations from the "Are we done with ImageNet" paper. The dataset is available under the Apache License 2.0 at \url{https://github.com/google-research/reassessed-imagenet}.

\paragraph{ImageNet-Hard}
ImageNet-Hard is a new benchmark featuring challenging images curated from various ImageNet validation datasets. It challenges state-of-the-art vision models as simply zooming in often fails to improve classification accuracy. The dataset is available under the MIT License at \url{https://github.com/taesiri/ZoomIsAllYouNeed}.

\begin{figure*}[t]
    \centering
    \includegraphics[width=\textwidth]{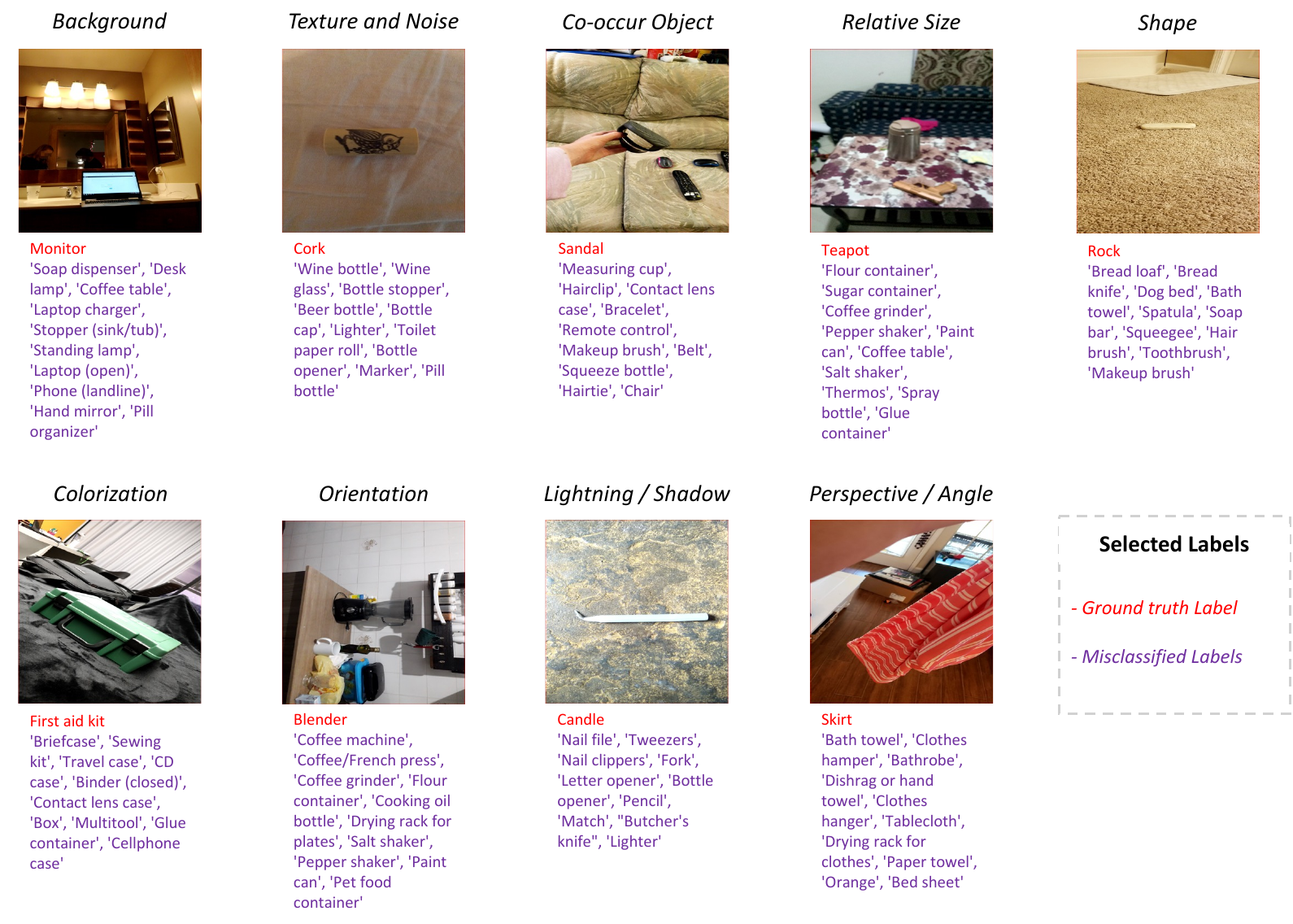}
    \caption{Examples of ground truth and misclassified labels after zero-shot classification. The \textcolor{red}{red} color shows the ground truth label and the \textcolor{mediumorchid}{purple} color shows 10 misclassified labels selected by our method.} \label{fig:misclassified}
\end{figure*}

\section{Disuccsion on Definition 3.1}
\label{sec:derivation}
Building on Section 3.2 and Definition 3.1, we explore how spurious biases learned in the vision encoder and the LLM can propagate into MLLMs. To analyze this, we represent the training data probabilities in the vision encoder and LLM of the MLLM using a spurious attribute $\mathbf{z}$, the core object $c$, and the core features $x_{\text{core}}$ and $y_{\text{core}}$ as follows.

\begin{align}
    &\textbf{In Vision Encoder: } p(\mathbf{z} | c, x_{\text{core}}) \gg p(\mathbf{z} | x_{\text{core}}) \label{eq:encoder}
\end{align}
\begin{align}
    &\textbf{In LLM: } p(\mathbf{z} | c, y_{\text{core}}) \gg p(\mathbf{z} | y_{\text{core}}) \label{eq:llm}
\end{align}

The conditional probability on the spurious attribute $\mathbf{z}$, given the core features and object, is:
\begin{align}
    p(\mathbf{z}|x_{\text{core}},y_{\text{core}},c)&=\frac{p(x_{\text{core}},y_{\text{core}}|\mathbf{z},c)p(\mathbf{z}|c)}{p(x_{\text{core}},y_{\text{core}}|c)}  \\
    &=\frac{p(x_{\text{core}}|\mathbf{z},c)p(y_{\text{core}}|\mathbf{z},c)p(\mathbf{z}|c)}{p(x_{\text{core}}|c)p(y_{\text{core}}|c)}\\
    &= \frac{p(\mathbf{z}|x_{\text{core}},c)p(\mathbf{z}|y_{\text{core}},c)p(\mathbf{z}|c)}{p(\mathbf{z}|c)p(\mathbf{z}|c)} \\
    &= \frac{p(\mathbf{z}|x_{\text{core}},c)p(\mathbf{z}|y_{\text{core}},c)}{p(\mathbf{z})}
\end{align}
Without considering the core object $c$, the conditional probability on the spurious attribute $\mathbf{z}$ is:
\begin{align}
    p(\mathbf{z}|x_{\text{core}},y_{\text{core}})&=\frac{p(x_{\text{core}},y_{\text{core}}|\mathbf{z})p(\mathbf{z})}{p(x_{\text{core}},y_{\text{core}})} \\
    &= \frac{p(x_{\text{core}}|\mathbf{z})p(y_{\text{core}}|\mathbf{z})p(\mathbf{z})}{p(x_{\text{core}},y_{\text{core}})}\\
    &= \frac{p(\mathbf{z}|x_{\text{core}})p(\mathbf{z}|y_{\text{core}})p(\mathbf{z})p(x_{\text{core}})p(y_{\text{core}})}{p(\mathbf{z})p(\mathbf{z})p(x_{\text{core}},y_{\text{core}})} \\
    &=\frac{p(\mathbf{z}|x_{\text{core}})p(\mathbf{z}|y_{\text{core}})}{p(\mathbf{z})}\cdot \frac{p(x_{\text{core}})p(y_{\text{core}})}{p(x_{\text{core}},y_{\text{core}})}\\
      &\approx \frac{p(\mathbf{z}|x_{\text{core}})p(\mathbf{z}|y_{\text{core}})}{p(\mathbf{z})}
\end{align}

By \eqref{eq:encoder} and \eqref{eq:llm}, we can get inequality \eqref{eq:multimodal} in the multimodal case. 
\begin{align}
p(\mathbf{z}|x_{\text{core}},c)p(\mathbf{z}|y_{\text{core}},c)&\gg p(\mathbf{z}|x_{\text{core}})p(\mathbf{z}|y_{\text{core}}) \\
\frac{p(\mathbf{z}|x_{\text{core}},c)p(\mathbf{z}|y_{\text{core}},c)}{p(\mathbf{z})}&\gg \frac{p(\mathbf{z}|x_{\text{core}})p(\mathbf{z}|y_{\text{core}})}{p(\mathbf{z})} \\
p(\mathbf{z}|x_{\text{core}},y_{\text{core}},c) &\gg p(\mathbf{z}|x_{\text{core}},y_{\text{core}}) 
\end{align}

We show that multimodal spurious biases can be exacerbated in the vision and language modalities. In this work, we assume \textit{the case that the spurious attribute $\mathbf{z}$ is shared between the vision encoder and LLMs}. Future work could explore more complex scenarios where the spurious attribute is not shared and investigate whether such cases further impact the robustness of MLLMs.

\begin{figure*}[t]
    \centering
    \includegraphics[width=\textwidth]{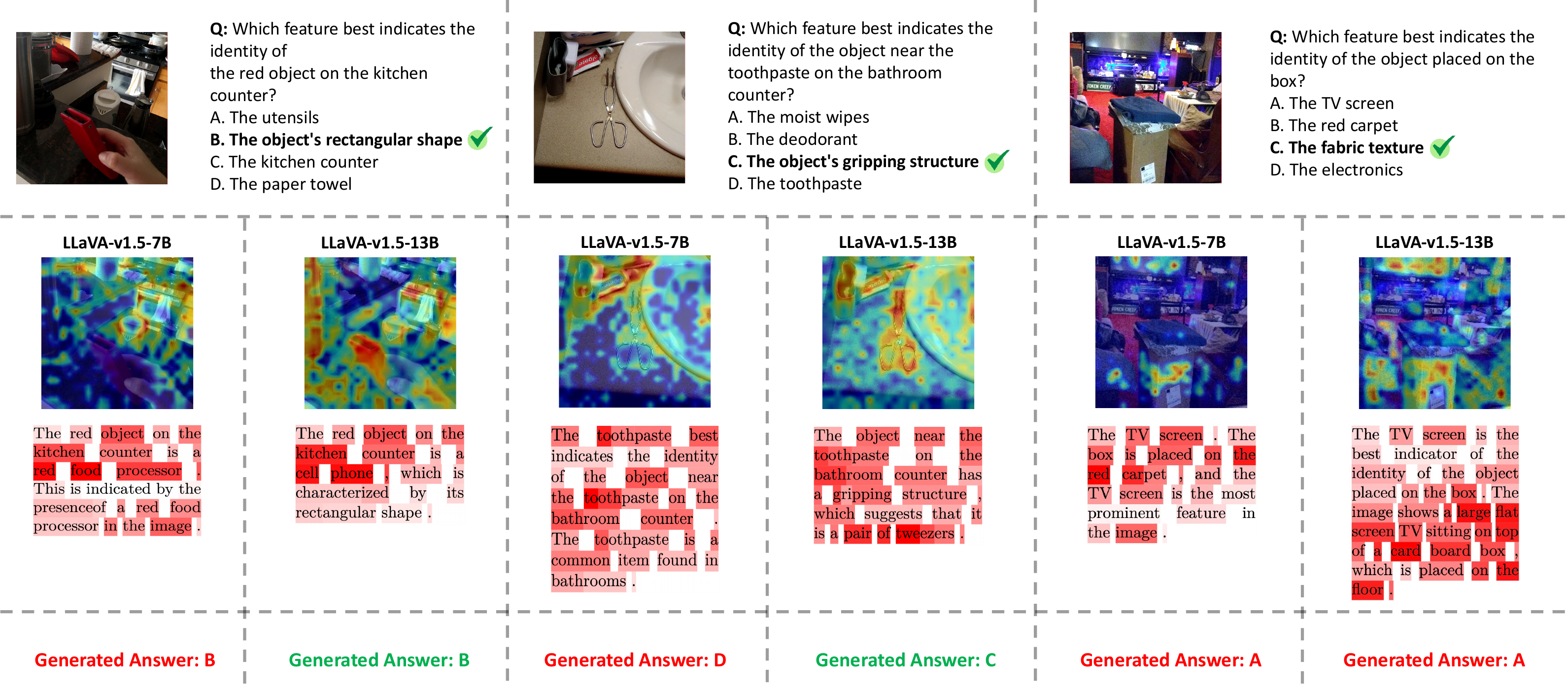} 
    \caption{GradCAM Visualization and Text Attention Heatmap of \textsc{MM-SpuBench} on LLaVA 7B and 13B. The \textit{anchor} objects in the three images are: a cellphone case, tweezers, and a full-sized towel.}
    \label{fig:gradcam}
\end{figure*}

\section{More Experimental Details}
\label{sec:more_expr}
\subsection{Zero-shot Classification}

In this section, we demonstrate that, in our designed image pre-selection stage, zero-shot classification using CLIP can effectively identify images with spurious correlations. This phenomenon has been previously observed and supported in several works. For instance, Table 4 in \cite{xue2024understanding} reports that the pretrained CLIP model, using the ViT-L/14@336px backbone, exhibited significantly lower worst-group accuracy (34.0\%) compared to average accuracy (88.5\%). However, the underperformance of CLIP models may also stem from factors such as label noise or annotation errors. To address this, we select images where CLIP’s true class prediction is not in the top-$k$ but appears in the top-$l$, allowing us to focus on potential spurious correlations.

We chose the CLIP model for zero-shot classification for several reasons. CLIP is widely adopted as a vision encoder in many of the most popular open-source MLLMs, such as LLaVA, which we use as baseline methods in Table~\ref{tab:open_model_acc}. Moreover, this paper highlights the potential for simple modality alignment methods to propagate spurious biases from the vision encoder to the broader MLLM framework. In Section 3.2, we provide evidence of how spurious biases observed in CLIP models can influence the entire MLLM pipeline, showing the relevance of addressing these biases at the source to enhance overall robustness.

As shown in Fig.~\ref{fig:misclassified}, misclassified labels often include spurious information unrelated to the core object but present in the image or contextually linked to it (e.g., ‘Cork’ vs. ‘Wine Bottle’). This supports the effectiveness of zero-shot classification in identifying potential spurious attributes. Zero-shot classification enables models to predict without prior exposure to specific examples, and analyzing misclassified labels helps reveal spurious attributes tied to the ground truth label. For instance, ‘Cork’ is frequently misclassified as ‘Wine Bottle’ or ‘Wine Glass,’ highlighting the model’s reliance on contextual cues over intrinsic features. Using \texttt{CLIP-ViT-L/14@336px}\footnote{https://huggingface.co/openai/clip-vit-large-patch14-336}, we identified specific spurious correlations that degrade model performance. For example, ‘Monitor’ was confused with ‘Soap Dispenser’ or ‘Desk Lamp’ due to background features, while ‘Sandal’ was misclassified as ‘Measuring Cup’ or ‘Hairclip’ due to similarities in shape and orientation.


\subsection{Prompt Engineering}
To ensure the effective generation and evaluation of questions for analyzing spurious correlations in images, we design the prompts for the MLLMs. In Fig.~\ref{fig:qa_prompt}, we created a system message prompt to guide the assistant in identifying spurious correlations and deriving core and spurious attributes. We then formulate multiple-choice questions that test a model's ability to distinguish these attributes. This ensures challenging and accurately reflective questions of spurious biases. 

For zero-shot evaluation on open-source models, as shown in Fig.~\ref{fig:zeroshot_prompt_open}, we directly ask the model to select the best answer. In contrast, for closed-source models, as illustrated in Fig.~\ref{fig:zeroshot_prompt_close}, we designed a straightforward prompt instructing the assistant to answer questions based on the provided image and four answer options, focusing on selecting the best answer and providing a brief explanation. Some open-source models may not consistently follow instructions and provide answers in the correct format. To address this, we do not explicitly require these models to include reasoning in their responses. However, for proprietary models, we ask them to first provide a brief explanation of their reasoning in one sentence before generating the final answer. This approach ensures that the models better comprehend and address the visual question-answering (VQA) task.

Additionally, we used the chain-of-thought prompt to enhance the assistant's reasoning capability by considering the type of spurious correlation provided in the benchmark and thinking step-by-step before choosing the best answer in Fig.~\ref{fig:zeroshot_prompt_close_cot}. In this work, we only use a simple strategy with two steps to do chain-of-thought reasoning, as we don't want to expose too much information to make the benchmark less challenging. However, in practice, it is recommended to use the concept information (core and spurious attributes) as mentioned in our benchmark to guide the MLLMs with better focus, as long as we determine the anchor object in the data. Our experiments serve as a motivating example for future research to develop improved strategies for enhancing the robustness of MLLMs to spurious biases.

\begin{table*}[ht]
\centering
\caption{Text Propmts of Tested Prompt Methods}
\label{tab:prompting_strategy}
\begin{tabular}{l|p{12cm}}
\toprule
\textbf{Strategy} & \textbf{Prompt} \\
\midrule
\textbf{Guiding} &
\textit{"You need to describe the image first. Pay attention to the details that confirm your answer. Be mindful of the unrelated attributes in the image. Finally, end your answer with one letter."} \\
\midrule
\textbf{No Bias} &
\textit{"Ensure that your answer is not affected by unimportant co-occurrence and spurious attributes. End your response with a one-letter answer."} \\
\midrule
\textbf{Explain} &
\textit{"To answer the question, you need to explain why a certain option identifies the type of the object in the image. Finally, end your response with one one-letter answer."} 
 \\
\bottomrule
\end{tabular}
\end{table*}

\subsection{Visualization Details}

In this section, we provide the experimental details of the visualizations and additional Grad-CAM visualizations and text attention heatmaps of \textsc{MM-SpuBench} on LLaVA 7B and 13B, as shown in Fig.~\ref{fig:gradcam}. For these visualizations, we prompted the MLLMs to reason through their generated answers using the template provided in Fig.~\ref{fig:reason_prompt}. We limited the generated outputs to a maximum of 50 tokens for clarity and excluded truncated sentences. 
For the Grad-CAM visualizations, we selected the last two layers of the LLaVA vision encoder to analyze the model’s visual focus and understand its interpretation of the input. For the text attention heatmaps, we first tokenized the generated reasoning and then computed attention values within the LLMs. These methods provide insights into the areas of focus in the MLLMs that lead to errors or failures in reasoning.


\begin{table*}[!ht]
\newcommand*{\SetGradientLimits}[2]{%
            \renewcommand*{\MinNumber}{#1}%
            \renewcommand*{\MaxNumber}{#2}%
        }

\centering
\caption{Zero-shot results of different open-source MLLMs on \textsc{MM-SpuBench} with varying prompting methods. All numbers are percentage accuracies. {\setlength{\fboxsep}{1pt}\colorbox{goodgreen!100}{Green}} color indicates higher accuracy, and {\setlength{\fboxsep}{1pt}\colorbox{goodred!100}{red}} color indicates lower accuracy. }
\label{tab:prompt_type_accuracy}

\begin{tabularx}{\textwidth}{llH{0}{100}H{0}{100}H{0}{100}H{0}{100}H{0}{100}H{0}{100}H{0}{100}H{0}{100}H{0}{100}H{0}{100}}
\toprule
\multicolumn{1}{c}{\textbf{Model}} & \multicolumn{1}{c}{\textbf{Prompt}} &   \multicolumn{1}{c}{\texttt{BG}} &
  \multicolumn{1}{c}{\texttt{TN}} &
  \multicolumn{1}{c}{\texttt{CO}} &
  \multicolumn{1}{c}{\texttt{RS}} &
  \multicolumn{1}{c}{\texttt{Col.}} &
  \multicolumn{1}{c}{\texttt{Ori.}} &
  \multicolumn{1}{c}{\texttt{LS}} &
  \multicolumn{1}{c}{\texttt{PA}} &
  \multicolumn{1}{c}{\texttt{Sha.}} &
  \multicolumn{1}{c}{\textbf{Average}} \\
\midrule

\multirow{4}{*}{ InternVL3-8B~\cite{zhu2025internvl3exploringadvancedtraining} }& Vanilla & 74.48 & 70.87 & 78.01 & 58.87 & 60.42 & 73.79 & 70.67 & 68.00 & 67.33 & 71.92\\& Guiding & 78.83 & 75.97 & 81.28 & 59.68 & 63.19 & 72.82 & 78.67 & 67.00 & 67.33 & 75.46\\& No Bias & 77.94 & 73.98 & 80.37 & 63.71 & 62.85 & 73.79 & 70.67 & 69.00 & 71.07 & 74.96\\& Explain & 75.05 & 71.54 & 78.14 & 60.48 & 61.46 & 74.76 & 76.00 & 67.00 & 69.33 & 72.67\\
\midrule
\multirow{4}{*}{ InternVL3-14B~\cite{zhu2025internvl3exploringadvancedtraining} }& Vanilla & 81.55 & 80.51 & 82.85 & 64.92 & 66.32 & 75.73 & 80.00 & 69.00 & 64.09 & 77.92\\& Guiding & 78.35 & 77.19 & 79.58 & 60.48 & 61.46 & 73.79 & 76.00 & 68.00 & 64.34 & 74.88\\& No Bias & 83.44 & 82.39 & 85.08 & 65.32 & 66.67 & 77.67 & 82.67 & 73.00 & 66.08 & 79.75\\& Explain & 80.61 & 79.62 & 82.59 & 62.10 & 63.89 & 75.73 & 78.67 & 67.00 & 63.84 & 76.96\\
\midrule
\multirow{4}{*}{ InternVL3-38B~\cite{zhu2025internvl3exploringadvancedtraining} }& Vanilla & 86.11 & 83.28 & 88.22 & 73.79 & 70.14 & 85.44 & 81.33 & 76.00 & 74.56 & 83.04\\& Guiding & 85.90 & 83.94 & 88.87 & 69.76 & 66.32 & 86.41 & 78.67 & 74.00 & 74.06 & 82.62\\& No Bias & 86.37 & 83.94 & 88.22 & 74.60 & 70.14 & 86.41 & 81.33 & 76.00 & 74.06 & 83.29\\& Explain & 86.06 & 83.94 & 87.43 & 73.39 & 69.44 & 86.41 & 80.00 & 74.00 & 75.06 & 82.96\\
\midrule
\multirow{4}{*}{Llama-3.2-VI-11B~\cite{meta2024llama32}}& Vanilla & 74.58 & 73.53 & 76.57 & 59.27 & 62.85 & 68.93 & 74.67 & 63.00 & 61.35 & 71.71\\& Guiding & 76.21 & 74.86 & 78.53 & 59.68 & 63.89 & 73.79 & 77.33 & 66.00 & 57.11 & 72.92\\& No Bias & 76.73 & 74.53 & 76.57 & 57.26 & 63.54 & 74.76 & 78.67 & 64.00 & 59.85 & 72.79\\& Explain & 77.73 & 74.53 & 79.71 & 60.89 & 64.24 & 79.61 & 78.67 & 69.00 & 61.85 & 74.25\\
\midrule
\multirow{4}{*}{LLaVA-v1.5-7B~\cite{liu2024visual} }& Vanilla & 41.19 & 41.31 & 39.53 & 33.47 & 34.38 & 40.78 & 38.67 & 38.00 & 35.16 & 39.54\\& Guiding & 41.25 & 41.31 & 39.66 & 33.87 & 34.03 & 39.81 & 40.00 & 38.00 & 35.66 & 39.62\\& No Bias & 40.04 & 40.31 & 37.43 & 34.27 & 34.03 & 38.83 & 40.00 & 35.00 & 33.92 & 38.38\\& Explain & 41.51 & 40.86 & 39.40 & 35.08 & 35.07 & 40.78 & 40.00 & 37.00 & 34.91 & 39.67\\
\midrule
\multirow{4}{*}{LLaVA-v1.5-13B~\cite{liu2024visual}}&  Vanilla & 63.31 & 62.57 & 63.48 & 39.52 & 46.18 & 55.34 & 56.00 & 50.00 & 47.63 & 59.08\\& Guiding & 59.75 & 57.25 & 59.95 & 39.52 & 45.14 & 49.51 & 50.67 & 52.00 & 45.64 & 55.71\\& No Bias & 63.89 & 61.46 & 64.40 & 38.31 & 47.57 & 55.34 & 53.33 & 54.00 & 46.63 & 59.25\\& Explain & 64.10 & 61.79 & 64.27 & 41.94 & 48.26 & 59.22 & 53.33 & 53.00 & 47.63 & 59.75\\
\midrule
\multirow{4}{*}{LLaVA-v1.6-mis-7B~\cite{liu2024visual}}& Vanilla & 40.88 & 41.20 & 41.36 & 34.27 & 39.24 & 39.81 & 40.00 & 41.00 & 34.41 & 39.96\\& Guiding & 40.57 & 38.65 & 41.62 & 34.68 & 36.46 & 35.92 & 45.33 & 33.00 & 33.17 & 39.00\\& No Bias & 40.15 & 39.87 & 40.05 & 34.27 & 41.32 & 41.75 & 41.33 & 40.00 & 33.92 & 39.33\\& Explain & 34.17 & 34.00 & 35.73 & 33.47 & 36.81 & 33.01 & 36.00 & 31.00 & 27.18 & 33.83\\
\midrule
\multirow{4}{*}{LLaVA-v1.6-vic-13B~\cite{liu2024visual} }&Vanilla & 21.75 & 20.16 & 22.77 & 15.73 & 26.39 & 21.36 & 18.67 & 17.00 & 13.97 & 20.75\\& Guiding & 25.31 & 24.36 & 25.52 & 22.98 & 23.96 & 25.24 & 22.67 & 26.00 & 21.95 & 24.67\\& No Bias & 52.46 & 52.60 & 54.32 & 41.94 & 46.88 & 43.69 & 52.00 & 41.00 & 41.40 & 50.58\\& Explain & 50.68 & 51.27 & 52.49 & 42.74 & 46.88 & 43.69 & 49.33 & 41.00 & 40.90 & 49.29\\
\midrule
\multirow{4}{*}{Qwen2-VL-7B~\cite{bai2023qwen} }&Vanilla & 73.95 & 69.44 & 74.21 & 52.02 & 56.25 & 69.90 & 73.33 & 61.00 & 55.36 & 69.04\\& Guiding & 73.27 & 68.44 & 73.95 & 53.23 & 56.60 & 71.84 & 73.33 & 63.00 & 61.35 & 69.17\\& No Bias & 71.59 & 67.55 & 72.25 & 48.39 & 52.78 & 68.93 & 68.00 & 64.00 & 56.61 & 67.04\\& Explain & 75.58 & 69.88 & 76.44 & 54.03 & 59.03 & 71.84 & 73.33 & 63.00 & 62.84 & 71.12\\
\midrule
\multirow{4}{*}{Qwen2.5-VL-7B~\cite{Qwen2.5-VL}  }& Vanilla & 71.59 & 67.55 & 73.43 & 53.63 & 58.33 & 73.79 & 74.67 & 56.00 & 61.85 & 68.38\\& Guiding & 63.31 & 58.47 & 64.92 & 50.00 & 52.08 & 54.37 & 72.00 & 53.00 & 56.11 & 60.42\\& No Bias & 71.59 & 67.88 & 72.51 & 56.45 & 59.72 & 70.87 & 77.33 & 59.00 & 64.59 & 68.79\\& Explain & 64.62 & 60.13 & 65.84 & 50.40 & 51.74 & 63.11 & 73.33 & 55.00 & 54.36 & 61.54\\
\bottomrule
\end{tabularx}
\end{table*}

\begin{figure*}[t]
\begin{AIbox}{Prompt template for QA Generation}

\TBlack{ {\bf System Message} } \\
You are a helpful assistant that analyze images.

\TBlack{ {\bf User Prompt} } \\
I will give you an
image: <image> \\
true label: ...
misclassified labels: ...

Spurious correlations are brittle associations learned by the models between non-essential spurious attributes of inputs and the corresponding core learning attributes in the training dataset. 

Based on the provided information
1. Figure out what kind of spurious correlations is performing in the given image. 
2. Based on the true label and the image, generate what are the core attributes of this true object label. Based on the misclassified labels and the image, generate what are the spurious attributes that are causing the misclassification.
3. Generate a multiple choice question based on the analysis to test the capability of a model whether it can identify the true label based on the spurious attributes. Among the choices, there should be only one correct answer related to the core attributes. Make the other choices as misleading as possible so that the model may fail on it.
4. Do not provide the true label or the core attributes of the main object in the question. Only use its visible spurious attributes or its spatial position in the image to refer to the object.

The max words for each attribute is \{max\_words\_per\_attribute\}. \\
The max number of core attributes is \{num\_core\_attributes\}. \\
The max number of spurious attributes is \{num\_spurious\_attributes\}. \\
For the generated multiple choice questions, the number of correct options is 1, and the number of wrong options is \{num\_wrong\_options\}. 

You should only respond in the format as described below: 

\TBlack{ {\bf Response Format} } \\
\textit{Explanation:} The explanation of the attributes. \\
\textit{Core Attributes:} The core attributes of the main object, must be visible in the image. \\
\textit{Spurious Attributes:} The spurious attributes in the image. \\
\textit{Spurious Correlation Type:} Should be from the 9 possible categories: Background; Texture and Noise; Co-occurring Objects; Relative Size; Colorization; Orientation; Lighting and Shadows; Perspective and Angle; Shape. Two at most. \\
\textit{Questions:} The question to ask about the image.\\
\textit{Choices:} The choices for the question, indexed by a single letter. \\
\textit{Answer:} The index of the correct answer, as a single letter.
\end{AIbox}
\caption{Prompt template for system message and response format for the QA generation with GPT-4V.}
\label{fig:qa_prompt}
\end{figure*}

\begin{figure*}[ht]
\begin{AIbox}{Prompt template for zero-shot evaluation}
\TBlack{ {\bf System Message} } \\
You are a helpful assistant that can answer question for an image. I will provide you 4 options. 
\\
\TBlack{ {\bf User Prompt} } \\
<image> \\
Here is the question: ... \\
Here are the choices: \\
A. ... \\
B. ... \\
C. ... \\
D. ... \\
\TBlack{ {\bf Response Format} } \\
\textit{Choice:} A single character from {A, B, C, D}.
\end{AIbox}
\caption{Prompt template for system message and response format for the zero-shot evaluation on open-sourced models.}
\label{fig:zeroshot_prompt_open}
\end{figure*}

\begin{figure*}[ht]
\begin{AIbox}{Prompt template for zero-shot evaluation}
\TBlack{ {\bf System Message} } \\
You are a helpful assistant that can answer question for an image. I will provide you 4 options. \\
\TBlack{ {\bf User Prompt} } \\
<image> \\
Here is the question: ... \\
Here are the choices: \\
A. ... \\
B. ... \\
C. ... \\
D. ... \\
\\
\TBlack{ {\bf Response Format} } \\
\textit{Explanation:} Explanation text in one sentence. \\
\textit{Choice:} A single character from {A, B, C, D}.
\end{AIbox}
\caption{Prompt template for system message and response format for the zero-shot evaluation on proprietary models.}
\label{fig:zeroshot_prompt_close}
\end{figure*}

\begin{figure*}[ht]
\begin{AIbox}{Prompt template for chain-of-thought evaluation}
\TBlack{ {\bf System Message} } \\
You are a helpful assistant that can answer question based on the image. Spurious correlations are brittle associations learned between non-essential spurious attributes of inputs and the corresponding core learning attributes. I will first provide you with the potential spurious correlation existing in the image. Then I will ask you a question with 4 options. Think step by step and then choose the best answer from the choices. \\
\TBlack{ {\bf Step 1} } \\
The type of spurious correlation occurs in the images are <type1> and <type2>. \\
<type1> occurs when the model .... One example is ... \\
<type1> occurs when the model .... One example is ... \\

\TBlack{ {\bf Step 2} } \\
<image> \\
Here is the question: ... \\
Here are the choices: \\
A. ... \\
B. ... \\
C. ... \\
D. ... \\
\\
\TBlack{ {\bf Response Format} } \\
\textit{Explanation:} Explanation text in one sentence. \\
\textit{Choice:} A single character from {A, B, C, D}.
\end{AIbox}
\caption{Prompt template for system message and response format for the chain-of-thought evaluation on proprietary models.}
\label{fig:zeroshot_prompt_close_cot}
\end{figure*}

\begin{figure*}[ht]
\begin{AIbox}{Assistant Prompt template for CGL calculation}
\TBlack{ {\bf System Message} } \\
You are a helpful assistant that can answer question based on the image.
\\
\TBlack{ {\bf User Prompt} } \\
<image> \\
What's the distinguishing attribute of the object? What is <object\_literal>? Answer in the format: I see [object\_attribute], so the mentioned object is of type: [object\_class].\\
\TBlack{ {\bf Assistant} } \\
I see <attribute\_text>, so the mentioned object is of type: <start\_of\_generation> \\

\end{AIbox}

\begin{AIbox}{User Prompt template for CGL calculation}
\TBlack{ {\bf System Message} } \\
You are a helpful assistant that can answer question based on the image.
\\
\TBlack{ {\bf User Prompt} } \\
<image> \\
This image shows <attribute\_text>. What is <object\_literal>? Answer in this format: This is an object of type: <object\_class>." \\
\TBlack{ {\bf Assistant} } \\
This is an object of type: <start\_of\_generation> \\

\end{AIbox}
\caption{Prompt template for the CGL evaluation on open-sourced models.}
\label{fig:cgp_prompt_open}
\end{figure*}

\begin{figure*}[ht]
\begin{AIbox}{Prompt template for reasoning in visualization}
\TBlack{ {\bf System Message} } \\
You are a helpful assistant that can answer questions based on the image. 
Provide a concise answer and explain your reasoning clearly.

\TBlack{ {\bf User Prompt} } \\
<image> \\
Here is the question: ... \\
Here are the choices: \\
A. ... \\
B. ... \\
C. ... \\
D. ... \\
\\
\TBlack{ {\bf Response Format} } \\
\textit{Explanation:} Provide a detailed explanation of your reasoning for selecting the choice. \\
\textit{Choice:} A single character from {A, B, C, D}.
\end{AIbox}
\caption{Prompt template for reasoning in Grad-CAM visualization on LLaVA.}
\label{fig:reason_prompt}
\end{figure*}

\begin{table*}[h]
    \centering
    \resizebox{\textwidth}{!}{%
    \begin{tabular}{>{\centering\arraybackslash}m{3cm} >{\centering\arraybackslash}m{6cm} >{\raggedright\arraybackslash}m{5cm} >{\centering\arraybackslash}m{1cm} >{\centering\arraybackslash}m{3.5cm}}
        \toprule
        \textbf{Image} & \textbf{Question} & \textbf{Choices} & \textbf{Answer} & \textbf{Type} \\
        \midrule
        \includegraphics[width=3cm, height=3cm, trim={0.5cm 0.5cm 0.5cm 0.5cm}, clip]{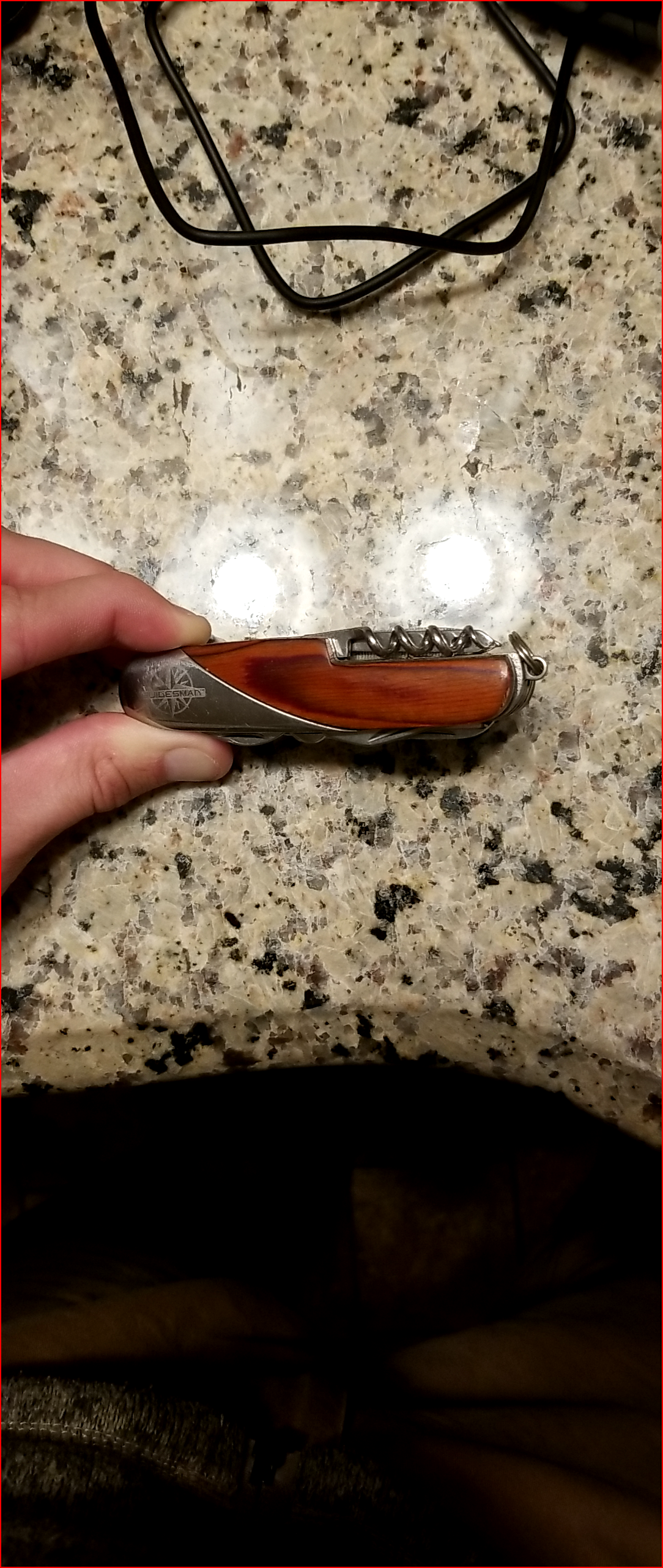} &
        Which feature best indicates the identity of the object \textcolor{red}{held in hand on the countertop}? &
        \begin{tabular}{@{}l@{}} \textcolor{red}{A. The countertop} \\ \textcolor{red}{B. The lighting} \\ \textcolor{red}{C. The metallic parts} \\ \textcolor{darkgreen}{D. The multiple tools}\end{tabular} &
        D &
        \begin{tabular}{c}Texture and Noise \\ Shape\end{tabular} \\
        \midrule
        \includegraphics[width=3cm, height=3cm, trim={0.5cm 0.5cm 0.5cm 0.5cm}, clip]{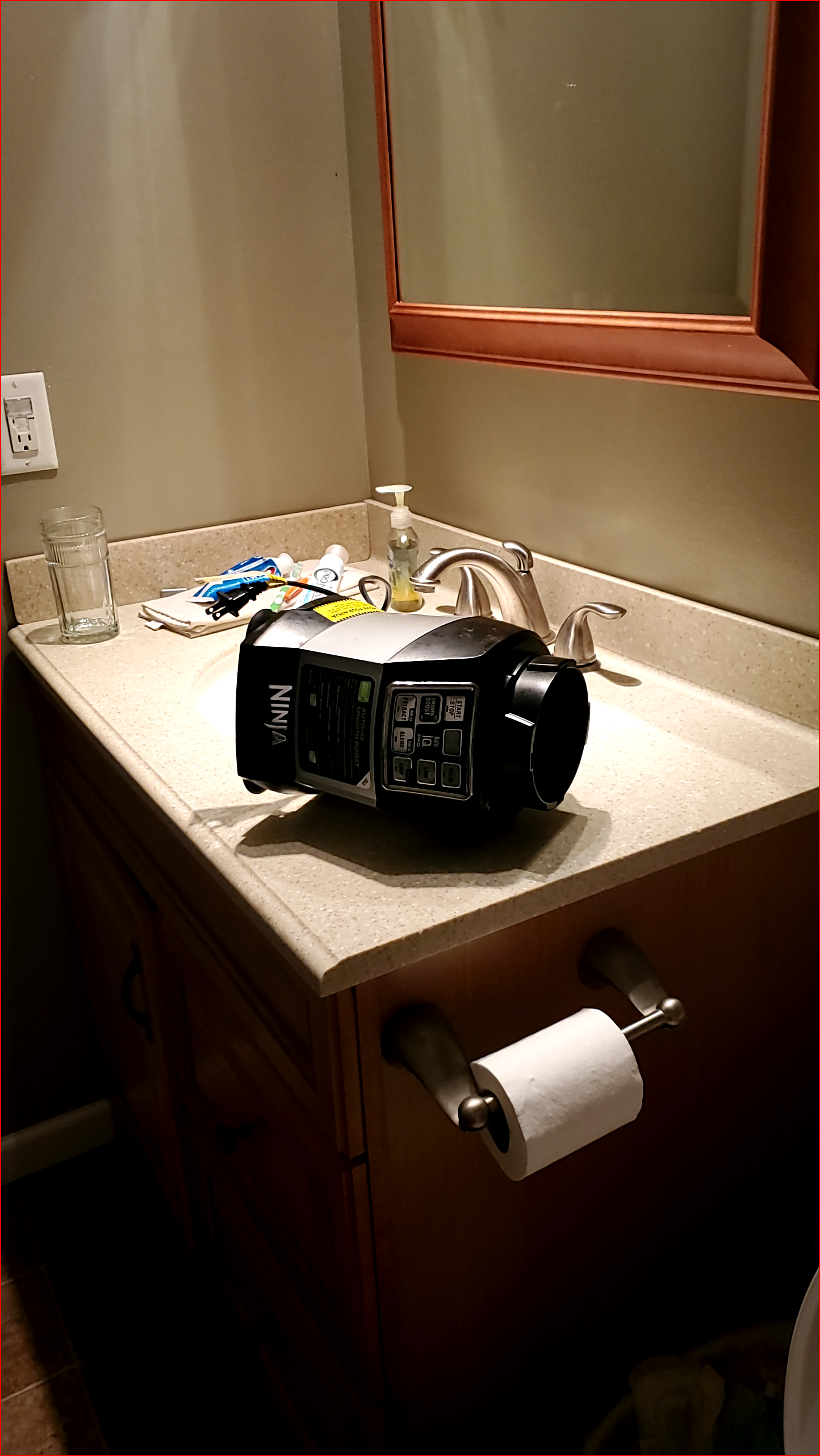} &
        Which feature best indicates the identity of the object that is \textcolor{red}{lying horizontally on the bathroom sink}? &
        \begin{tabular}{@{}l@{}} \textcolor{darkgreen}{A. The object's control panel} \\ \textcolor{red}{B. The bathroom mirror} \\ \textcolor{red}{C. The toothbrushes} \\ \textcolor{red}{D. The soap dispenser}\end{tabular} &
        A &
        \begin{tabular}{c}Background \\ Orientation\end{tabular} \\
        \midrule
        \includegraphics[width=3cm, height=3cm, trim={0.5cm 0.5cm 0.5cm 0.5cm}, clip]{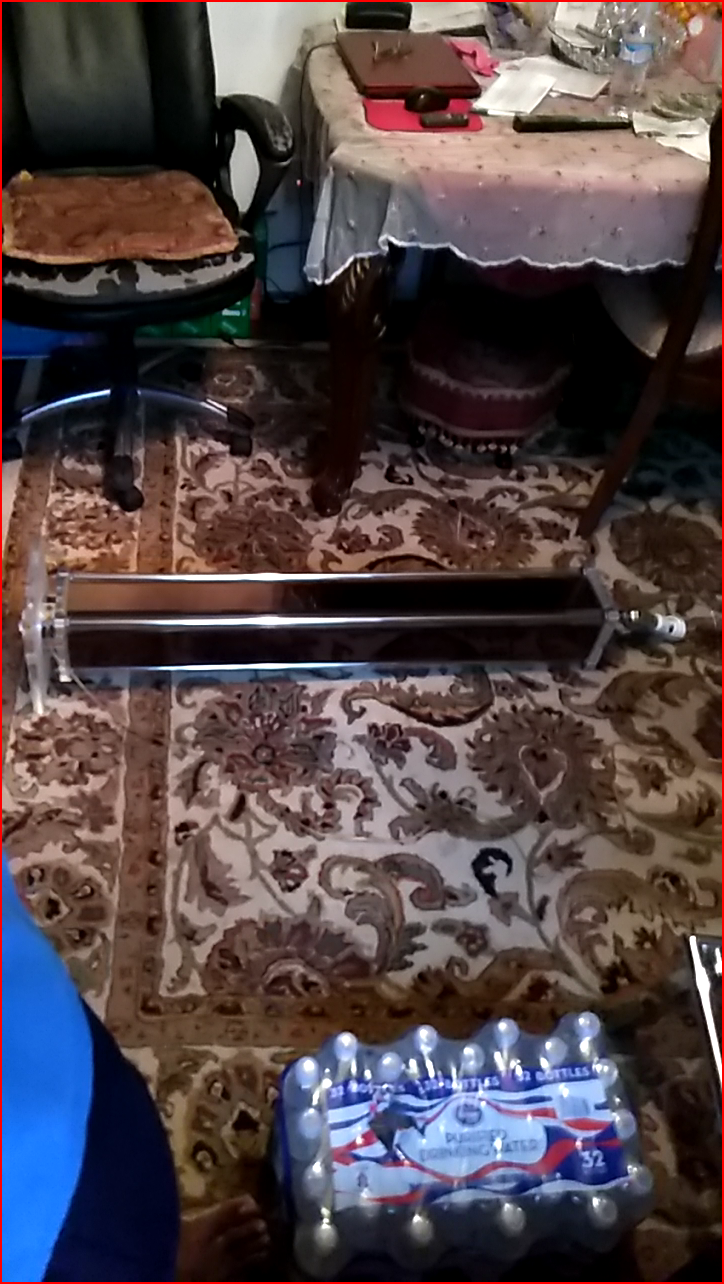} &
        Which feature best indicates the identity of the object that is \textcolor{red}{lying horizontally on the floor}? &
        \begin{tabular}{@{}l@{}} \textcolor{red}{A. Office chair} \\ \textcolor{red}{B. Water bottles} \\ \textcolor{red}{C. Table} \\ \textcolor{darkgreen}{D. Light bulb socket}\end{tabular} &
        D &
        \begin{tabular}{c}Orientation \\ Co-occurring Objects\end{tabular} \\
        \midrule
        \includegraphics[width=3cm, height=3cm, trim={0.5cm 0.5cm 0.5cm 0.5cm}, clip]{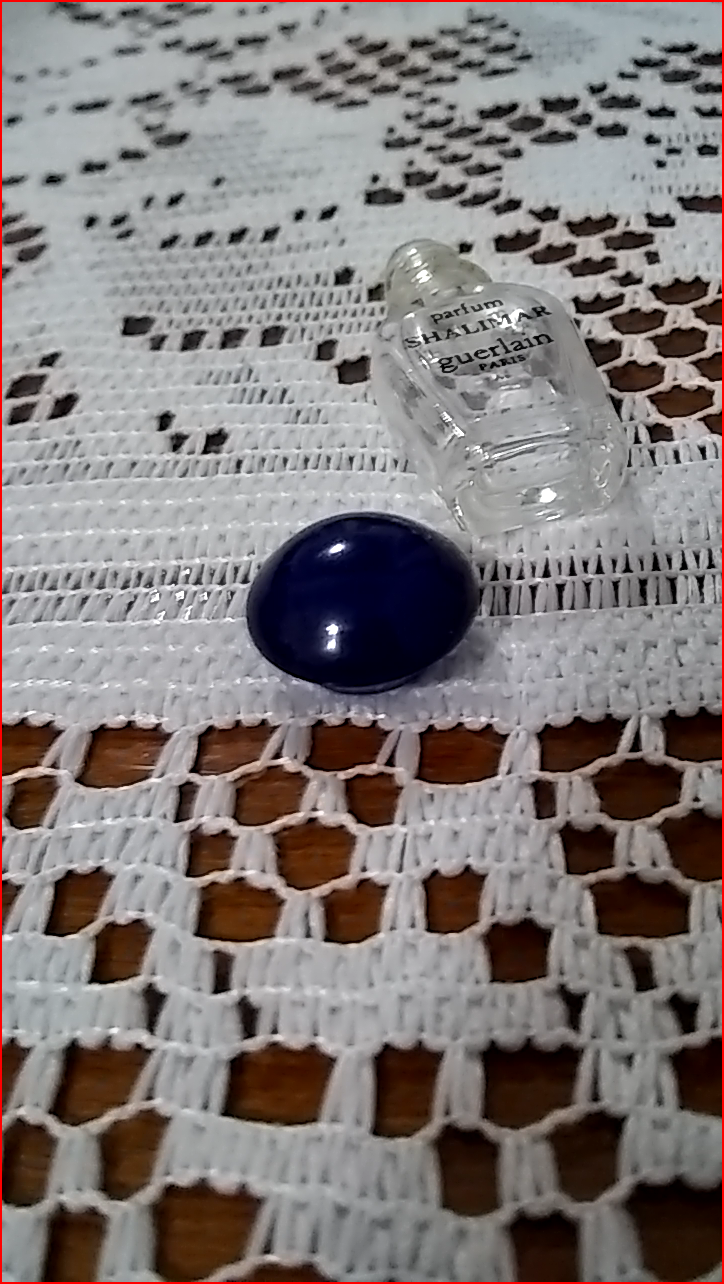} &
        Which feature best indicates the identity of \textcolor{red}{the small cylindrical object next to the small bottle}? &
        \begin{tabular}{@{}l@{}} \textcolor{red}{A. Its small size} \\ \textcolor{darkgreen}{B. Its cylindrical shape} \\ \textcolor{red}{C. The lace tablecloth} \\ \textcolor{red}{D. The nearby bottle}\end{tabular} &
        B &
        \begin{tabular}{c}Co-occurring Objects \\ Relative Size\end{tabular} \\
        \bottomrule
    \end{tabular}%
    }
    \caption{More Data instances in \textsc{MM-SpuBench}. Images are cropped and resized to fit in the table. \textcolor{red}{Red} denotes the spurious attributes and \textcolor{darkgreen}{green} denotes the core attributes.}
    \label{tab:data_instance_part1}
\end{table*}

\begin{table*}[t]
    \centering
    \resizebox{\textwidth}{!}{%
    \begin{tabular}{>{\centering\arraybackslash}m{3cm} >{\centering\arraybackslash}m{6cm} >{\raggedright\arraybackslash}m{5cm} >{\centering\arraybackslash}m{1cm} >{\centering\arraybackslash}m{3.5cm}}
        \toprule
        \textbf{Image} & \textbf{Question} & \textbf{Choices} & \textbf{Answer} & \textbf{Type} \\
        \midrule
        \includegraphics[width=3cm, height=3cm, trim={0.5cm 0.5cm 0.5cm 0.5cm}, clip]{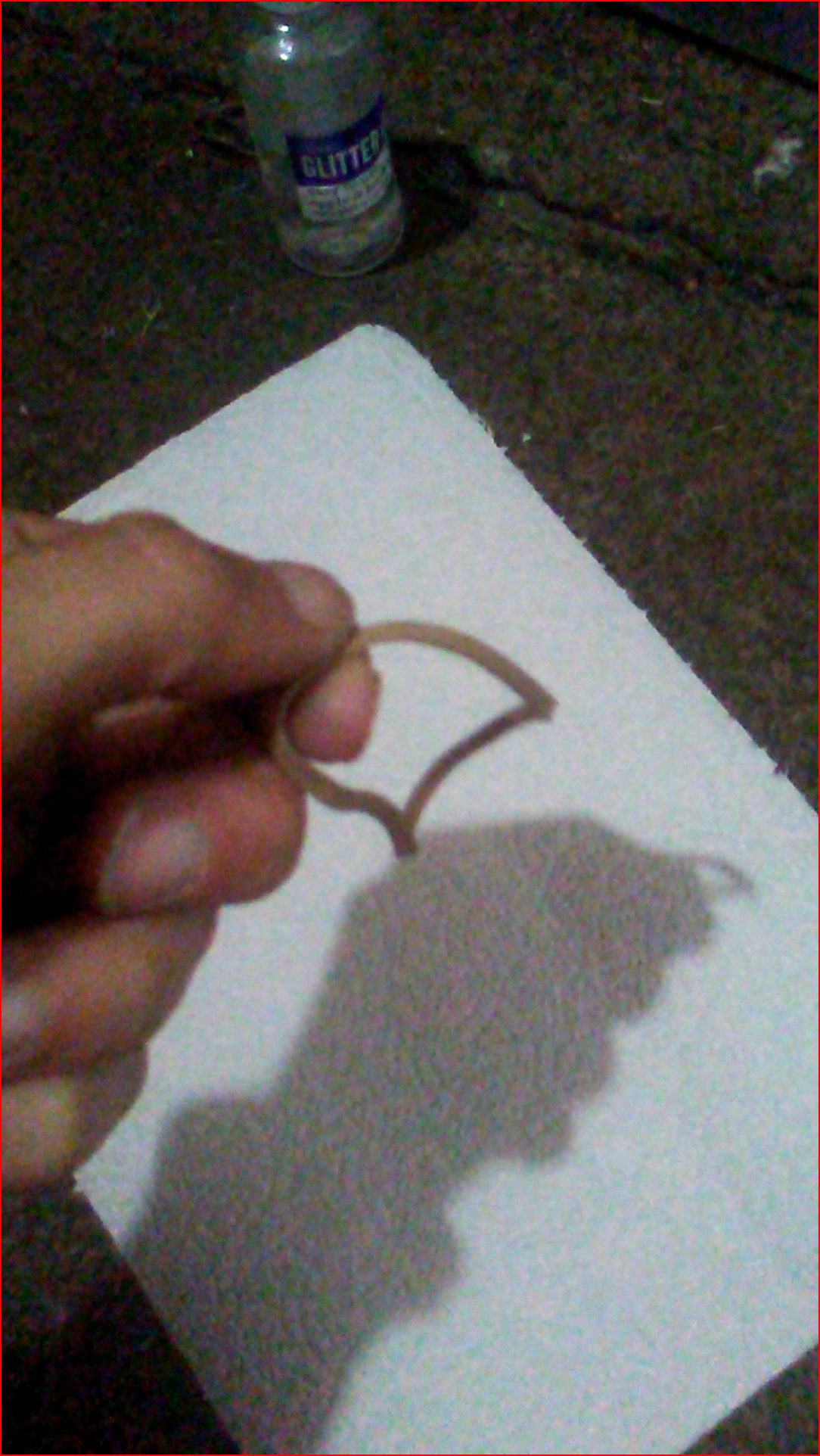} &
        Which feature best indicates the identity of the object \textcolor{red}{being held}? &
        \begin{tabular}{@{}l@{}} \textcolor{red}{A. The glitter bottle} \\ \textcolor{darkgreen}{B. The object's circular shape} \\ \textcolor{red}{C. The shadow} \\ \textcolor{red}{D. The surface texture}\end{tabular} &
        B &
        \begin{tabular}{c}Background \\ Lighting and Shadows\end{tabular} \\
        \midrule
        \includegraphics[width=3cm, height=3cm, trim={0.5cm 0.5cm 0.5cm 0.5cm}, clip]{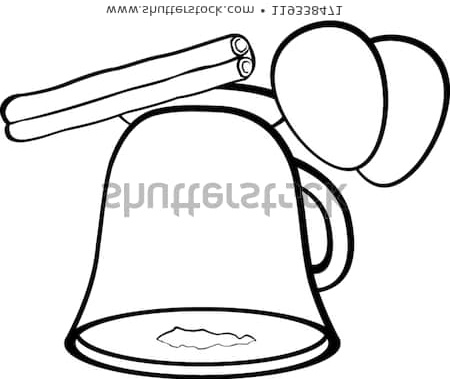} &
        Which feature best indicates the identity of the object that is upside down \textcolor{red}{with two sticks on top}? &
        \begin{tabular}{@{}l@{}} \textcolor{red}{A. Circular shapes} \\ \textcolor{red}{B. The shark-like form} \\ \textcolor{red}{C. The cinnamon sticks} \\ \textcolor{darkgreen}{D. The curved handle}\end{tabular} &
        D &
        \begin{tabular}{c}Orientation \\ Shape\end{tabular} \\
        \midrule
        \includegraphics[width=3cm, height=3cm, trim={0.5cm 0.5cm 0.5cm 0.5cm}, clip]{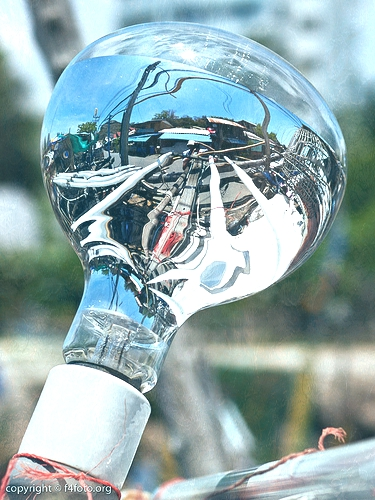} &
        Which feature best indicates the identity of the object with a \textcolor{red}{distorted reflection}? &
        \begin{tabular}{@{}l@{}} \textcolor{darkgreen}{A. The bulb shape} \\ \textcolor{red}{B. The background buildings} \\ \textcolor{red}{C. The outdoor setting} \\ \textcolor{red}{D. The lightening}\end{tabular} &
        A &
        \begin{tabular}{c}Background \\ Perspective and Angle\end{tabular} \\
        \midrule
        \includegraphics[width=3cm, height=3cm, trim={0.5cm 0.5cm 0.5cm 0.5cm}, clip]{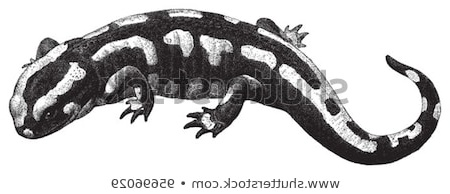} &
        Which feature best indicates the identity of the animal \textcolor{red}{with black and white patterns}? &
        \begin{tabular}{@{}l@{}} \textcolor{red}{A. Small size} \\ \textcolor{darkgreen}{B. Dog-like head shape} \\ \textcolor{red}{C. Black and white pattern} \\ \textcolor{red}{D. Elongated body}\end{tabular} &
        B &
        \begin{tabular}{c}Colorization \\ Orientation\end{tabular} \\
        \midrule
        \includegraphics[width=3cm, height=3cm, trim={0.5cm 0.5cm 0.5cm 0.5cm}, clip]{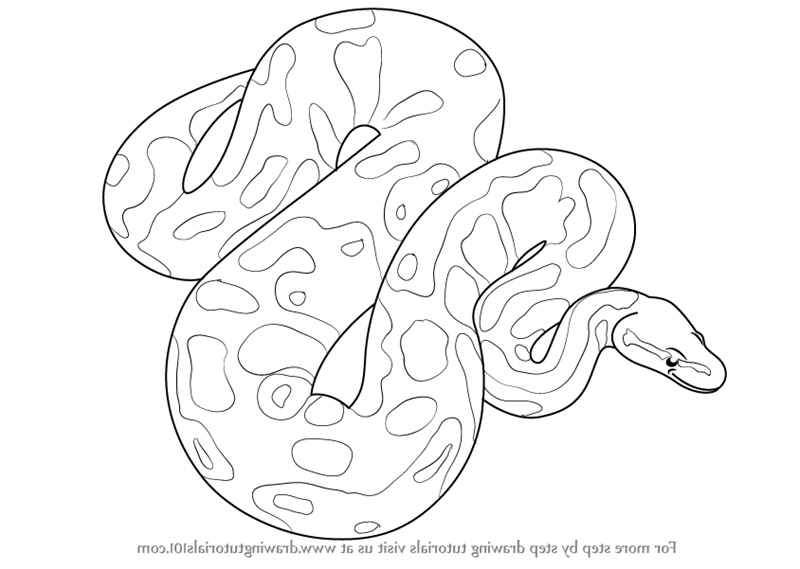} &
        Which feature best indicates the identity of the \textcolor{red}{coiled object} in the drawing? &
        \begin{tabular}{@{}l@{}} \textcolor{red}{A. The patterns on the object} \\ \textcolor{darkgreen}{B. The serpentine body} \\ \textcolor{red}{C. The outline drawing} \\ \textcolor{red}{D. The coiled shape}\end{tabular} &
        B &
        \begin{tabular}{c}Shape \\ Texture and Noise\end{tabular} \\
        \bottomrule
    \end{tabular}%
    }
    \caption{More data instances in \textsc{MM-SpuBench}. Images are cropped and resized to fit in the table. \textcolor{red}{Red} denotes the spurious attributes and \textcolor{darkgreen}{green} denotes the core attributes.}
    \label{tab:data_instance_part2}
\end{table*}

\end{document}